\documentclass[pdflatex,sn-mathphys-num,iicol]{sn-jnl}

\usepackage[table]{xcolor}
\usepackage{graphicx}%
\usepackage{multirow}%
\usepackage{amsmath,amssymb,amsfonts}%
\usepackage{amsthm}%
\usepackage{mathrsfs}%
\usepackage[title]{appendix}%
\usepackage{xcolor}%
\usepackage{textcomp}%
\usepackage{manyfoot}%
\usepackage{booktabs}%
\usepackage{algorithm}%
\usepackage{algorithmicx}%
\usepackage{algpseudocode}%
\usepackage{listings}%
\usepackage{bm}
\usepackage{makecell}
\usepackage{multirow}
\usepackage{amsfonts}
\usepackage{pifont}

\theoremstyle{thmstyleone}%
\theoremstyle{thmstyletwo}%

\theoremstyle{thmstylethree}%

\begin{document}

\title[Article Title]{T2LDM++: A Self-Conditioned Representation Guided Diffusion Model for Realistic Text-to-LiDAR Scene Generation}


\author[1]{\fnm{Wentao} \sur{Qu}} \email {quwentao@njust.edu.cn} 

\author*[2]{\fnm{Qi} \sur{Zhang}}\email{qizhang@cityu.edu.mo}


\author[3]{\fnm{Chenxu} \sur{Wang}} 

\author[4]{\fnm{Guofeng} \sur{Mei}} 

\author[1]{\fnm{Yongfei} \sur{Liu}} 

\author[5]{\fnm{Xiaoshui} \sur{Huang}} 

\author[6]{\fnm{Gim Hee} \sur{ Lee}} 

\author*[1]{\fnm{Liang} \sur{Xiao}} \email{xiaoliang@mail.njust.edu.cn}

\affil[1]{\orgdiv{School of Computer Science and Engineering}, \orgname{Nanjing University of Science and Technology}, \orgaddress{\city{Nanjing}, \postcode{210094}, \country{China}}}

\affil[2]{\orgdiv{Faculty of Data Science}, \orgname{City University of Macau}, \orgaddress{\city{Macau}, \postcode{999078}, \country{China}}}

\affil[3]{\orgdiv{School of Computer Science and Technology}, \orgname{Beijing Institute of Technology}, \orgaddress{\city{Beijing}, \postcode{100081}, \country{China}}}

\affil[4]{\orgname{Fondazione Bruno Kessler}, \orgaddress{\city{Trento}, \postcode{38123}, \country{Italy}}}

\affil[5]{\orgdiv{School of Public Health}, \orgname{Shanghai Jiao Tong University}, \orgaddress{\city{Shanghai}, \postcode{200030}, \country{China}}}

\affil[6]{\orgdiv{School of Computing}, \orgname{National University of Singapore}, \orgaddress{\city{Singapore}, \postcode{119077}, \country{Singapore}}}





\abstract{Recent progress in Text-to-Image generation benefits from large-scale Text-Image pairs. However, the scarcity of Text-LiDAR pairs often causes over-smoothed scenes and limited controllability. In this paper, we rethink the limitations of Text-LiDAR generation task, focusing on alleviating insufficient training priors and constructing controllable Text-LiDAR data. We propose a \textbf{T}ext-\textbf{to}-\textbf{L}iDAR \textbf{D}iffusion \textbf{M}odel for LiDAR scene generation, T2LDM++, with a Self-Conditioned Representation Guidance (SCRG). Specifically, to alleviate object over-smoothing, SCRG employs a Guidance Network (GN) to provide reconstruction-based soft supervision to the Denoising Network (DN).
This enables DN to learn geometry-aware representations through reconstruction guidance, leading to more accurate denoising in DDPMs. Meanwhile, through analysis and design, SCRG exhibits more effective and lightweight, while decoupled in inference, avoiding computational overhead. Furthermore, we construct two high-quality Text-LiDAR benchmarks ($>$100K samples) using a generalized strategy of geometric annotations, along with a controllability metric. Moreover, a directional position prior is designed to mitigate street distortion, further improving scene fidelity. Additionally, T2LDM++ supports multiple conditions, including (Semantic, Box, BEV, Camera)-to-LiDAR, Sparse-to-Dense, and Dense-to-Sparse generation, by learning a control encoder via frozen DN. With effective prior modeling and high-quality Text-LiDAR benchmarks, T2LDM++ can generate realistic LiDAR scenes with rich geometric details in unconditional and conditional settings. \href{https://github.com/QWTforGithub/T2LDM_v2}{The code has been published}.}

%
%
%




\maketitle

\section{Introduction}\label{sec1}

\begin{figure*}[htp]
	\centering
	\includegraphics[width=0.99\textwidth]{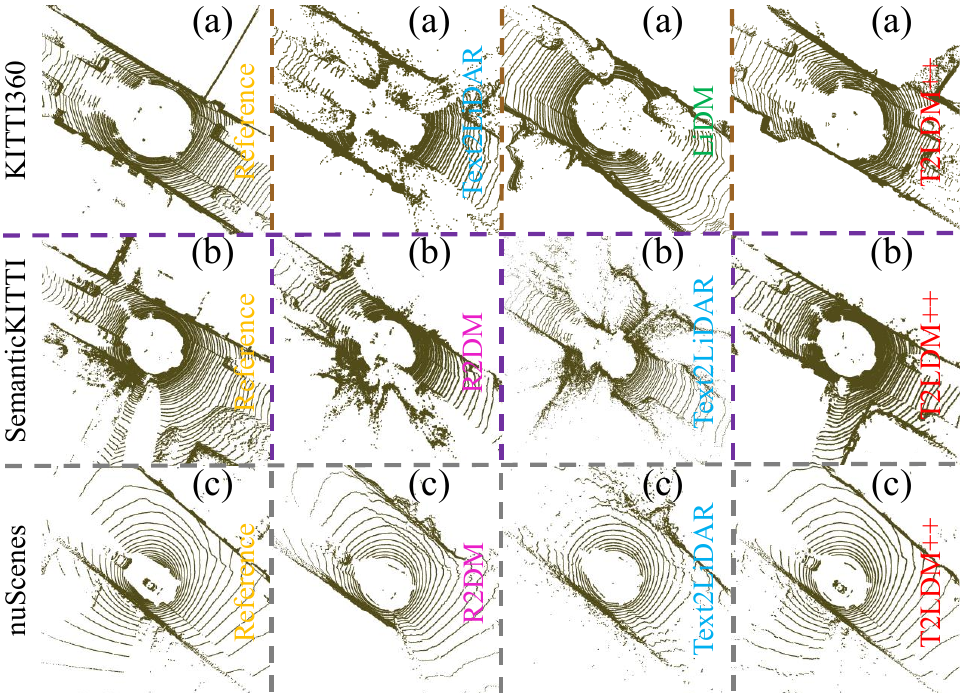}

	\caption{Results of existing LiDAR generation methods. Due to insufficient training priors, they struggle to produce detail-rich scene objects (a, b and c). This is more pronounced on small-scale datasets (SemanticKITTI, the LiDAR samples $<$ 25K, b) and sparse LiDAR data (nuScenes, 32-beam, c). In contrast, T2LDM++ can generate realistic and usable scenes.}
	\label{fig1}
	
\end{figure*}

LiDAR captures geometric structures and spatial layouts of 3D scenes by sensing the surrounding environment. This provides an essential sensing foundation and data support for downstream 3D tasks such as autonomous driving \cite{mao20233d, qu2025end}, AR/VR \cite{wang2023multi, behari2025blurred}, and robotics \cite{roldao20223d, qu2026robust}. However, collecting LiDAR scene data remains costly, especially when covering diverse layouts and rare weather conditions \cite{mei2024unsupervised, wu2024text2lidar, qu2026self}. This significantly limits the performance improvement and generalization ability of data-driven 3D perception models. Therefore, synthesizing realistic, diverse, and controllable LiDAR scenes has become increasingly crucial.

\begin{figure*}[htp]
	\centering
	\includegraphics[width=0.99\textwidth]{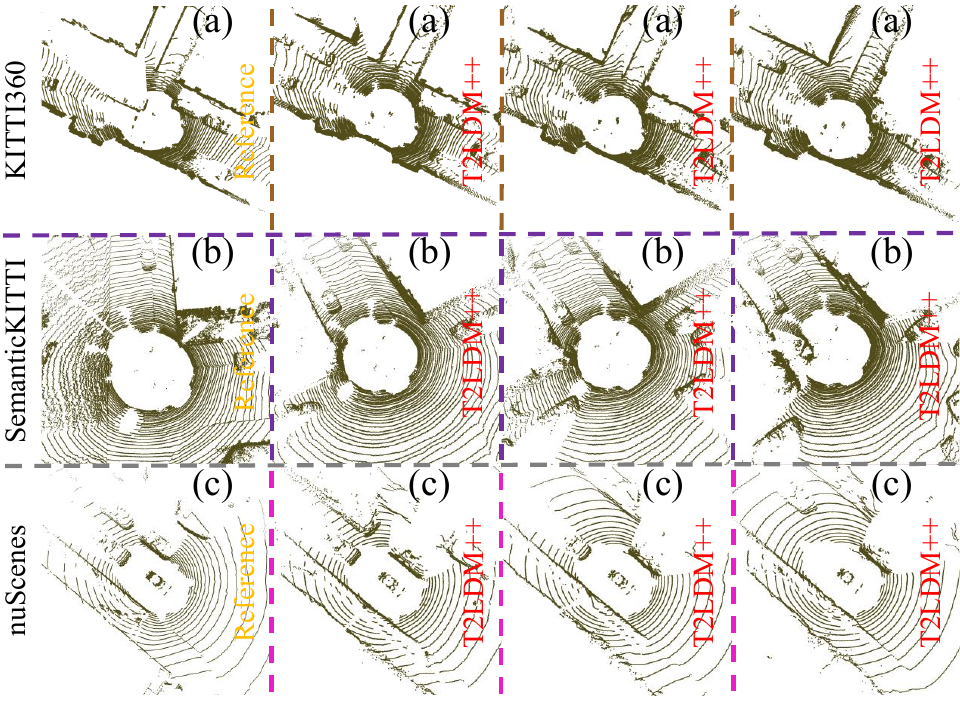}
	\caption{Visualization of multiple variants generated from the same scene for T2LDM++. Unlike existing methods that lack priors  and produce homogeneous results, T2LDM++ can generate diverse object distributions for the same scene. This demonstrates that T2LDM++ effectively captures scene geometry, enabling structurally consistent generation.}
	\label{fig2}
	
\end{figure*}

In recent years, significant progress has been made in Text-to-Image generation \cite{rombach2022high, saharia2022photorealistic, schuhmann2022laion, coyo700m2023, zhang2023adding}. This  mainly benefits from two factors: 1) Natural language provides convenient and controllable semantic guidance for generation tasks \cite{rombach2022high, saharia2022photorealistic}. 2) Large-scale Text-Image pairs offer strong alignment supervision for cross-modal generation \cite{radford2021learning, schuhmann2022laion, coyo700m2023}. Based on these, some methods can be even trained on over 100M Text-Image pairs \cite{ramesh2022hierarchical, saharia2022photorealistic, rombach2022high} to effectively synthesize realistic, diverse, and controllable content from natural language descriptions.

Inspired by these advances, some studies introduce text-guided generation for 3D scene customization \cite{wu2024text2lidar}. Similar to Text-to-Image synthesis, these methods build on generative frameworks like DDPMs \cite{ho2020denoising} by injecting text semantics into conditional diffusion models trained on existing LiDAR datasets \cite{behley2019semantickitti, caesar2020nuscenes, liao2022kitti}.


Unfortunately, unlike the easily collected largescale Text-Image pairs (e.g., from the open Internet \cite{radford2021learning, coyo700m2023}), acquiring 3D LiDAR scene data remains highly labor-intensive. Furthermore, provisioning accurate text annotations for such complex spatial data demands substantial manual effort. 
Consequently, high-quality and structurally diverse Text-LiDAR pairs are extremely scarce. For instance, the nuScenes benchmark \cite{caesar2020nuscenes} contains fewer than 35,000 pairs. This data scarcity severely impedes the effective training of generative models, leading to over-smoothed (see Fig.~\ref{fig1}) and homogeneous (see Fig.~\ref{fig2}) LiDAR scenes that lack fine-grained object details. This limitation is particularly pronounced in DDPMs, which rely heavily on large-scale training priors \cite{nakashima2024lidar, wu2024text2lidar}. Compounding this issue, while high-quality text descriptions are crucial for precise conditional generation, existing annotations are often rigidly structured due to high collection costs, failing to capture the natural variation of human language \cite{qu2026self}. Moreover, the lack of dedicated evaluation metrics hinders the assessment of generation controllability, making it difficult to quantitatively measure the cross-modal alignment between text prompts and synthesized LiDAR scenes.


To overcome these limitations, we rethink the current Text-to-LiDAR generation task by focusing on two key dimensions: 1) \textit{Enhancing training priors to alleviate object over-smoothing in synthesized scenes.} 2) \textit{Constructing high-quality, controllable Text-LiDAR benchmarks to mitigate data scarcity while enabling robust controllability evaluation.}





In this paper, we propose a \textbf{T}ext-\textbf{to}-\textbf{L}iDAR \textbf{D}iffusion \textbf{M}odel, named T2LDM++. Inspired by injecting regularization into DDPMs through representation learning \cite{li2023self, yu2024representation}, T2LDM++ introduces a Self-Conditioned Representation Guidance (SCRG). The key idea is to \textit{improve accurate denoising from a reconstruction perspective (Sec.~\ref{sec41}), because the model in DDPMs achieves the distribution matching by reconstructing the noise target under a reconstruction loss.} Specifically, SCRG employs a Guidance Network (GN) to provide the soft supervision with noise-adaptive reconstruction features for the Denoising Network (DN). Under this regularization, DN can continuously capture geometry-aware representations in denoising learning, accurately reconstructing the noise target. Therefore, this enables the sampling trajectory to perceive real geometric structures  (Fig.~\ref{fig8}(d)), generating realistic scenes with rich object details. Meanwhile, SCRG can be decoupled in inference, reducing the computational overhead and avoiding the information leakage. Furthermore, we conduct a systematic analysis (see Sec.~\ref{sec4}), revealing the key mechanism behind the effectiveness of SCRG. Based on these insights, we redesign the architecture, making SCRG more effective and lightweight. 


Meanwhile, we construct two high-quality Text-LiDAR benchmarks, T2nuScenes++ and T2SemanticKITTI, containing over 100K Text-LiDAR pairs. They are constructed from geometric annotations (3D Boxes and Semantic Labels) using simple geometric rules. This strategy offers several advantages. First, geometric annotations enable more precise text descriptions than manual annotation. Second, this can generalize to any dataset with geometric annotations. Third, this enables controllability evaluation using related task models. Finally, a large number of Text-LiDAR pairs can be produced by combining different types of text descriptions within a single scene. Actually, Text-to-LiDAR generation presents a more flexible and convenient paradigm than conditioning on geometric annotations. To the best of our knowledge, this is the first work to investigate producing text descriptions from geometric annotations, encouraging more researchers to explore text-guided 3D scene generation.  



Furthermore, we observe that flattening LiDAR data into the Range Map (RM) via spherical projection may introduce \textit{directional confusion} for the model.  Inspired by Circular Position Encodings \cite{lai2023spherical, su2024roformer} providing periodic angular position priors for Transformers in recognition tasks, we design a Directional Position Encoding (DPE) for generation tasks. This provides the real directional priors for T2LDM++ by defining angular coordinates in RM, further improving the fidelity of generated scenes. 

Additionally, by training a control Encoder through freezing DN, T2LDM++ can be extended to multiple conditional generation tasks. This is also the first attempt to apply \textit{a non-latent ControlNet \cite{zhang2023adding} for 3D generation}.

Our key contributions can be summarized as:

\begin{itemize}
	\item We propose a Text-to-LiDAR Diffusion Model, T2LDM++, with a Self-Conditioned Representation Guidance. This can guide the sampling trajectory to approximate real geometric structures, generating scenes with detailed objects.
	\item We construct two high-quality Text-LiDAR benchmarks using a generalized geometry-annotation-based strategy, accompanied by controllability metrics.
	\item We design a Directional Position Encoding, leveraging pixel-level angular priors to mitigate street distortion, further improving scene fidelity.
	\item Extensive experiments demonstrate that T2LDM++ can generate realistic LiDAR scenes with rich object details in unconditional and conditional generation.
\end{itemize}

A conference version of this work was accepted at CVPR 2026 \cite{qu2026self}. In this journal version, we extend the conference paper with the following contributions:

\begin{itemize}
	\item \textbf{Analysis and Optimization of SCRG.} Although SCRG has demonstrated effectiveness, the underlying mechanisms and architectural designs remain insufficiently explored. First, using a  segmentation linear probe, we reveal the mechanism that \textit{GN enhances the geometry-aware representation of DN}. Then, we demonstrate that \textit{SCRG effectively regularizes the denoising trajectory}, enabling DN to better approximate the reconstruction-consistent representation (the $\bm{x_0}$ features captured by GN) in training and inference via feature similarity analysis. Finally, building upon these insights, we redesign the SCRG framework, including the training strategy and the network architecture.

	\item \textbf{Extension of Benchmarks.} Text-LiDAR pairs constructed from 3D Boxes in nuScenes \cite{caesar2020nuscenes} remain limited in scale ($<$35K samples) \cite{qu2026self}. Meanwhile, this strategy remains underexplored for other geometric annotations, such as Semantic Labels. Therefore, we first extend T2nuScenes to over 100K samples, resulting in T2nuScenes++, demonstrating the effectiveness of associating a single scene with diverse types of text descriptions. Subsequently, by leveraging Semantic Labels, T2SemanticKITTI with over 100K Text-LiDAR pairs is constructed based on SemanticKITTI \cite{behley2019semantickitti}, further showing the generalizability of this strategy.
    
	\item \textbf{Multi-condition Expansion.} We introduce \textit{(Box, BEV, Camera)-to-LiDAR generation}, further validating extensibility of T2LDM++ using a non-latent ControlNet. 

    \item \textbf{Zero-Shot Text-LiDAR Generation.} In fact, text descriptions are derived from geometric annotations. This means that trained Box-to-LiDAR and Semantic-to-LiDAR T2LDM++ can achieve Zero-Shot Text-to-LiDAR generation. This insight provides an effective pipeline for Text-to-LiDAR generation.

\end{itemize}

To the best of our knowledge, \textit{T2LDM++ supports the most diverse conditions for LiDAR scene generation}.



\section{Related Works}\label{sec2}

\subsection{LiDAR Scene Generation}
LiDAR data provides an effective representation of real-world environments, enabling precise description of scene structures and object distributions. However, complex data acquisition, labor-intensive annotation processes, and rare weather conditions make high-quality and diverse LiDAR data difficult to obtain  \cite{caccia2019deep, zyrianov2022learning, nakashima2024lidar, wu2024text2lidar, wu2025weathergen}. Some methods try to synthesize realistic scenes based on existing LiDAR data through physics-based simulation \cite{manivasagam2020lidarsim, hahner2021fog, teufel2022simulating, yang2024realistic}. They typically model LiDAR physics based on optical scattering and laser propagation principles, simulating signal attenuation, backscattering, and measurement noise \cite{li2020happens, kilic2025lidar}. Although these methods partially simulate the physical imaging process, directly synthesizing realistic scenes from existing LiDAR scenes struggles to produce structurally rich and diverse layouts. This is because they rely on high-quality LiDAR data as the simulation basis,  limiting result diversity. Benefiting from the strong data-driven capability of deep learning, researchers have explored neural networks to generate LiDAR scenes with diverse layouts. \cite{caccia2019deep} is the first work to introduce deep generative models for LiDAR scene generation. This directly takes point clouds as input, generating LiDAR scenes using VAEs \cite{kingma2013auto} and GANs \cite{goodfellow2014generative}. However, the irregularity of point clouds makes models struggle to capture fine geometric details, leading to suboptimal generation quality. To address this issue, LiDARGen \cite{zyrianov2022learning} projects LiDAR data into Range Map \cite{milioto2019rangenet++} to obtain a regular representation, significantly improving generation quality using score-based models \cite{hyvarinen2005estimation}. Subsequently, building on this effective approach, R2DM \cite{nakashima2024lidar} further improves generation quality by adopting a stronger score-based model, DDPMs \cite{ho2020denoising}. Furthermore, LiDM \cite{ran2024towards} leverages latent diffusion \cite{rombach2022high} to achieve promising results under various conditional settings.

Although existing methods have achieved impressive LiDAR scene generation, insufficient training priors often lead to over-smoothed and homogeneous scenes, limiting the applicability. This becomes particularly evident in DDPMs, requiring large amounts of training priors \cite{wang2023patch, zhu2025domainstudio}. In this paper, we propose a Self-Conditioned Representation Guidance (SCRG), encouraging DN to learn geometry-aware representations in the progressive denoising process via geometry-aware regularization. This guides the sampling trajectory toward the reconstruction-consistent representation, improving object details and structural fidelity in generated scenes.


\subsection{Text-Guided Generation}
Natural language provides flexible semantic guidance for scene generation. Benefiting from the availability of large-scale Text–Image pairs \cite{schuhmann2022laion, coyo700m2023}, many methods can generate high-quality and diverse images aligned with given natural language descriptions, such as DALL·E2 \cite{ramesh2022hierarchical}, Imagen \cite{saharia2022photorealistic}, and Stable Diffusion \cite{rombach2022high}. Inspired by these advances, some researchers attempt to introduce text guidance into 3D generation. Early methods typically leverage Text-Image priors to bridge the gap between text and point clouds, enabling object-level Text-to-Point Cloud generation  \cite{nichol2022point, poole2022dreamfusion, lin2023magic3d}. However, the modality gap between images and point clouds limits the effectiveness of transferring Text-Image priors to 3D geometry. To overcome this, several works attempt to directly generate object-level 3D data from text prompts \cite{sanghi2022clip, jun2023shap, wu2023sketch}.
Recently, some methods have begun to explore Text-to-LiDAR scene generation, achieving promising preliminary results \cite{wu2024text2lidar}.

Some attempts have demonstrated the potential of Text-to-LiDAR generation, but the lack of high-quality Text-LiDAR pairs hinders further progress. In this paper, we construct two Text-LiDAR benchmarks. These benchmarks contain over 100K samples and provide controllable evaluation metrics. Meanwhile, the construction of Text-LiDAR pairs relies solely on geometric annotations, allowing this strategy to generalize to any dataset with the same type of annotations.

\subsection{Representation Learning for DDPMs}
\label{sec23}

“What I can not create, I do not understand.”

\quad\quad\quad\quad\quad\quad\quad\; ---Richard P. Feynman, 1988 

A thorough understanding of a system lies in revealing the underlying mechanisms, enabling generating new instances. In natural language processing, the success of T5 \cite{raffel2020exploring} and GPTs \cite{radford2019language} demonstrates understanding (representation) and generation can be unified. Inspired by this, researchers attempt to unify them in the vision domain. Variational Autoencoders (VAEs) \cite{kingma2013auto} are among the first deep learning frameworks to unify representation and generation. They employ an encoder to compress the input into a latent representation, then a decoder achieves the reconstruction to learn the underlying data distribution. Subsequently, some works leverage DDPMs \cite{ho2020denoising} as representation learners, achieving promising results in downstream tasks \cite{preechakul2022diffusion, wei2023diffusion, mittal2023diffusion, xiang2023denoising, yang2023diffusion, chen2024deconstructing}. Then, a natural question arises: \textit{Can representation learning be leveraged to improve generation quality?} Classifier Guidance \cite{dhariwal2021diffusion} steers the sampling trajectory in inference using a noise classifier, significantly improving the generation quality and controllability. Meanwhile, RCG \cite{li2024return} pretrains a latent representation that aligns the noise distribution with the image distribution. RCG uses this as conditional guidance to enhance unconditional generation. However, multi-stage training requires significant computational cost. Benefiting from large-scale knowledge priors, REPA \cite{yu2024representation} leverages pretrained self-supervised models \cite{oquab2023dinov2} to regularize the internal representations of DDPMs. This feeds clean images into the pretrained model to extract priors, employing contrastive learning to enhance the representation ability of DDPMs, significantly accelerating training convergence and improving generation performance. Furthermore, RAEs \cite{zheng2025diffusion} replace the VAE Encoder in latent diffusion with a pretrained representation prior \cite{caron2021emerging, oquab2023dinov2}, achieving effective the generation performance improvement.

Benefiting from large-scale available data in NLP and 2D vision, powerful pretrained priors provide strong representation support for generative models. However, due to the high cost of collection, the 3D domain lacks abundant high-quality data, struggling to obtain strong self-supervised priors. To address the above problem, inspired by improving generative models with representation learning \cite{leng2025repae}, we propose a regularization strategy that requires neither multi-stage training nor additional training data. This employs a lightweight Guidance Network (GN) to provide reconstruction-based soft supervision, encouraging the Denoising Network (DN) to learn geometry-aware representations in the denoising process, guiding the sampling trajectory toward the reconstruction-consistent representation.

\begin{figure*}[htp]
	\centering
	\includegraphics[width=0.99\textwidth]{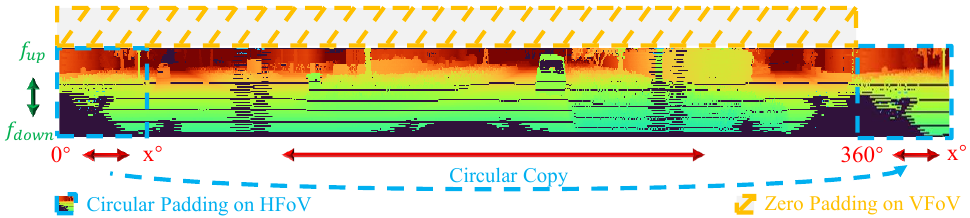}
	\caption{The padding approach of CircularConv2D. Applying circular padding in Decoder leads to over-smoothed reconstruction, as upsampling propagates the coupling between the left and right boundaries of reconstruction feature maps, making representations internally similar within features, diminishing local details.}
	\label{fig3}
	
\end{figure*}

\section{T2LDM++}\label{sec3}

\subsection{Generation Process}
\label{sec31}

\textbf{Input Representation.} T2LDM++ adopts Range Maps (RM) as the input representation \cite{nakashima2024lidar, wu2024text2lidar, ran2024towards}, due to the more regular structure than point clouds and the (partially) reversible transformation (LiDAR $\rightarrow$ RM $\rightarrow$ LiDAR). RM represents the global layout of LiDAR by projecting 3D coordinates onto a spherical surface (see Fig.~\ref{fig5}(a)$\rightarrow$(b)). The rows and columns correspond to the \textbf{H}orizontal ($0^\circ$-$360^\circ$, HFoV) and \textbf{V}ertical ($f_{down}$-$f_{up}$, VFoV) \textbf{F}ields \textbf{o}f \textbf{V}iew in the LiDAR space.  This projects $\bm{p}_i=(x,y,z)$ to $(u, v)$ via  \cite{milioto2019rangenet++}:

\begin{figure}[htp]
	\centering
	\includegraphics[width=0.48\textwidth]{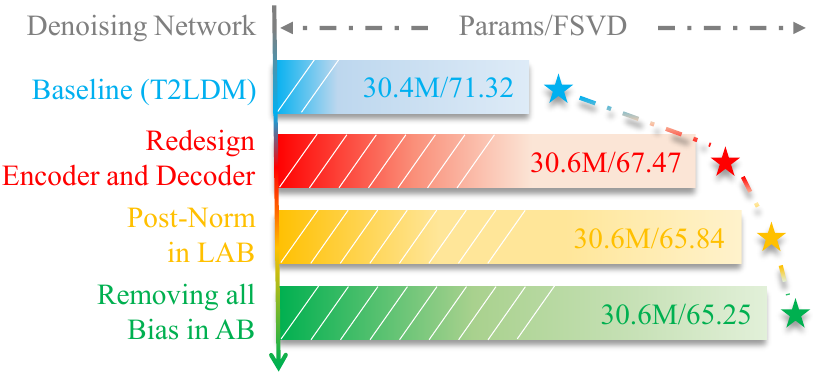}
	\caption{Through careful tuning, the backbone (DN, equipped with DPE, Sec.~\ref{sec33}) achieves a significant performance improvement over the previous version \cite{qu2026self}.}
	\label{fig4}
	
\end{figure}

\begin{equation}
	\begin{split}
		\label{f311}
		u = \frac{1}{2}[1-atan2(y,x)\pi^{-1}]W, \quad\;\;\; \\
		v = [1-(arcsin(zr^{-1})+f_{up})f^{-1}]H, 
	\end{split}
\end{equation}
where $(H, W)$ means the height and width of RM. Meanwhile, $r=||\bm{p}_i||^2$ represents the physical distance of $\bm{p}_i$ from the LiDAR sensor to the scene. Furthermore, $atan2(y,x)$ denotes the horizontal rotation angle (azimuth angle) of the point $(x,y)$.

Meanwhile, to enhance geometric representation, the depth $r$ and intensity $I$  are jointly used as pixel values of $\mathrm{RM} \in \mathbb{R}^{H \times W \times 2}$ \cite{nakashima2024lidar, wu2024text2lidar, qu2026self}.

\textbf{Text-to-LiDAR Generation.} A LiDAR point cloud is first projected onto RM ($\bm{x_0} \sim \mathcal{P}_{RM}$) via spherical projection. Then, given a text condition $\bm{c} \sim \mathcal{P}_{text}$ obtained from a Text Encoder \cite{raffel2020exploring, radford2021learning} and a prior noise $\bm{x_T} \sim \mathcal{P}_{noise}$, conditional DDPMs bridge $\mathcal{P}_{RM}$ and $\mathcal{P}_{noise}$ via: a predefined diffusion process $q$ that gradually perturbs $\bm{x_0}$ until $\bm{x_T}$, a trainable generation process $p_\theta$ that slowly cleans $\bm{x_T}$ until $\bm{x'_0}$ conditioned on $\bm{c}$. 

In the training process, the fitting objective of conditional DDPMs is:

\begin{equation}
	\begin{split}
		\label{f312}
		L(\theta) =
		\mathbb{E}_{\bm{\epsilon} \sim \mathcal{N}(0,I)}||\bm{v} - v_\theta(\bm{x_t},t,\bm{c})||^2, 
	\end{split}
\end{equation}
where the target $\bm{v}$ means a linear combination of $\bm{\epsilon}$ or $\bm{x_0}$ \cite{qu2026self}. 


In inference, $\bm{x_T} \sim \mathcal{P}_{noise}$ is iteratively converted back to $\bm{x'_0} \sim \mathcal{P}_{RM}$ by trained $v_\theta$. 

Finally, we can transform $\bm{x'_0}$ to the 3D coordinates to generate the LiDAR scene using the inverse of Eq.~\ref{f311} \cite{milioto2019rangenet++}.

\subsection{Enhanced Denoising Backbone} 
\label{sec32}

In this section, we introduce a redesigned backbone (Denoising Network, DN), further improving the baseline of T2LDM++.

\textbf{Encoder-Decoder Redesign.} Inspired by the success of Text-to-Image task, T2LDM \cite{qu2026self} follows the U-Net architecture of Stable Diffusion \cite{rombach2022high}. Unlike Rectangular Images (RI), RM, as a Circular Image (CI), requires convolution to model circular continuity along HFoV (left-right wrapping, see Fig.~\ref{fig3}). Therefore, T2LDM employs CircularConv2D \cite{stearns2024curvecloudnet} to model the geometric topology in RM, introducing a circular inductive bias along HFoV via circular padding. However, this may hinder the recovery of high-frequency details in Decoder, leading to strong coupling at feature boundaries (see Fig.~\ref{fig3}). Meanwhile, as the resolution increases (upsampling), this effect further propagates, leading to over-smoothed feature reconstruction. Therefore, we redefine the roles of Encoder and Decoder in DN:

\begin{itemize}
	\item Encoder: Feature compression, employing CircularConv2D to capture the geometric topology.
	\item Decoder: Feature reconstruction, employing Conv2D to reconstruct high-frequency details.
\end{itemize}


\textbf{Normalization in Attention Block.} T2LDM \cite{qu2026self} employs two types of attention blocks: Traditional Attention Block (TAB) \cite{vaswani2017attention} and Linear Attention Block (LAB) \cite{qu2026self}. Following empirical practices \cite{vaswani2017attention}, TAB typically adopts pre-norm for optimal performance. However, LAB employs a kernel-based approximation of attention \cite{choromanski2020rethinking}. The performance is highly dependent on the feature distribution. Pre-norm early normalizing the feature distribution may reduce magnitude differences across tokens, potentially weakening attention discrimination. In contrast, post-norm applies normalization after the projection layer, allowing attention to operate on relatively unnormalized features, better preserving magnitude variations in LAB.

\textbf{Bias in Attention Block.} Furthermore, we observe that bias in Attention Block (AB) plays a less critical role in geometric modeling within RM.  This is because geometric relationships are primarily determined by relative structural patterns rather than absolute value offsets \cite{lowe2004distinctive, qi2017pointnet, phan2018dgcnn}. 

Finally, the adjustments and the performance changes are illustrated in Fig.~\ref{fig4}.


\subsection{Directional Position Encoding} 
\label{sec33}

Although CircularConv2D effectively captures circular continuity along HFoV, the window-based unidirectional operation globally perceives RM as RI rather than CI (see Fig.~\ref{fig5}). This often hinders the model from capturing the correct spatial arrangement of objects in scenes, leading to \textit{directional confusion}. Meanwhile, since the projection typically starts from the street center, this effect is most pronounced as street distortion. 

In this paper, we design a Directional Position Encoding (DPE), to address \textit{directional confusion} in RM. By encoding the HFoV and VFoV angular coordinates, DPE injects directional priors, enabling the model to correctly perceive the object arrangement in RM. 

Specifically, given the internal feature $\bm{x} \in \mathbb{R}^{b \times c \times h \times w}$ from $v_\theta$, DPE first maps each pixel $(i,j)$ to the angular coordinates $(\bm{\theta},\bm{\phi})$ via:

\begin{equation}
	\begin{split}
		\label{f331}
		\bm{\theta} = 2\pi-[(2\pi-0)*(i+0.5)/w], \quad\;\; \\
		\bm{\phi} = f_{up}-[(f_{up}-f_{down})*(j+0.5)/h]. 
	\end{split}
\end{equation}

Then, the multi-level Fourier expansion models multi-scale directional priors:

\begin{figure}[htp]
	\centering
	\includegraphics[width=0.48\textwidth]{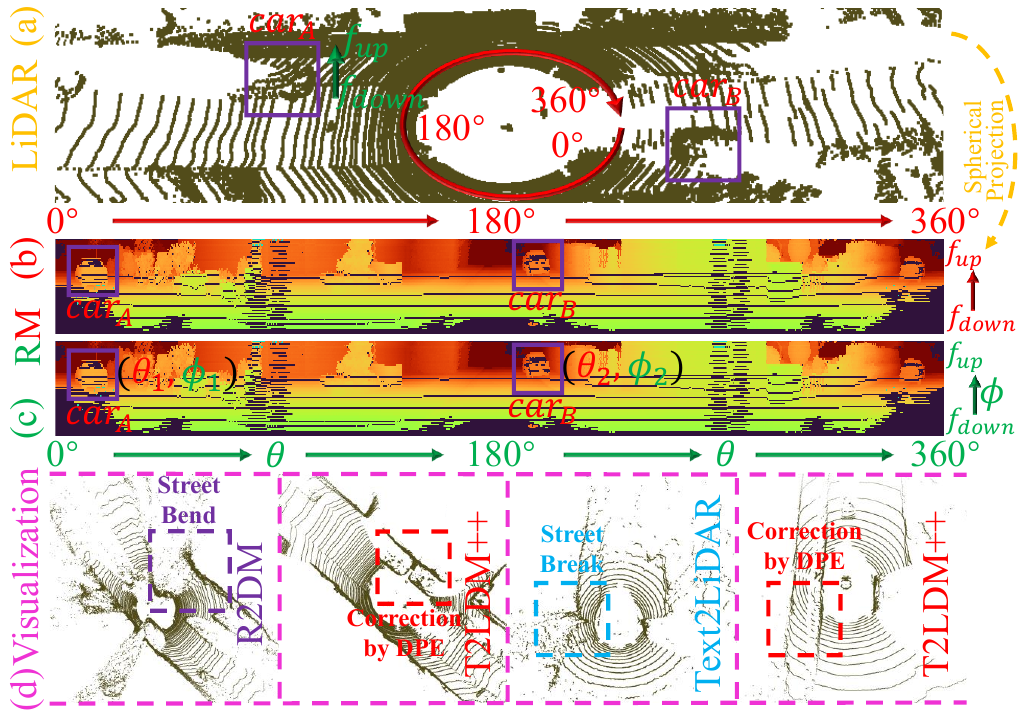}
	\caption{(a) In the LiDAR space, $car_A$ is at \textit{\textbf{the front-right}} of $car_B$. (b) However, window-based operations may perceive $car_A$ as being to \textit{\textbf{the left}} of $car_B$ in RM. (c) DPE explicitly provides the horizontal angle $\theta$ and the vertical angle $\phi$ priors in RM, enabling the model to correctly perceive true orientations of objects in the scene. This enables the model to clearly understand the relative spatial relationship between $car_A$ and $car_B$, improving scene fidelity. (d) Existing methods lacking directional priors may lead to bend or broken street structures. In contrast, T2LDM++ can generate realistic ones via the correction of DPE.}
	\label{fig5}
	
\end{figure}

\begin{equation}
	\begin{split}
		\label{f332}
		\mathrm{DPE}(\bm{\theta}, \bm{\phi}) = 
		Fourier^{K}(\bm{\theta}, \bm{\phi}),
	\end{split}
\end{equation}
where $K$ means the number of Fourier expansion terms. Meanwhile, $\bigoplus$ denotes the element-wise addition. Furthermore, $Fourier^{K}(\bm{\theta}, \bm{\phi})=\bigoplus_{k=0}^{K-1}[\sin(2^{k}\bm{\theta}),\cos(2^{k}\bm{\theta}),\sin(2^{k}\bm{\phi}),\cos(2^{k}\bm{\phi})]$.

Furthermore, unlike the fixed bias used in recognition tasks \cite{lai2023spherical}, we adopt a learnable gating to improve diversity in generation tasks: 

\begin{equation}
	\begin{split}
		\label{f333}
		\bm{x'}=\bm{x}+\alpha*\mathrm{DPE}(\bm{\theta}, \bm{\phi}).
	\end{split}
\end{equation}

\begin{figure*}[htp]
	\centering
	\includegraphics[width=0.99\textwidth]{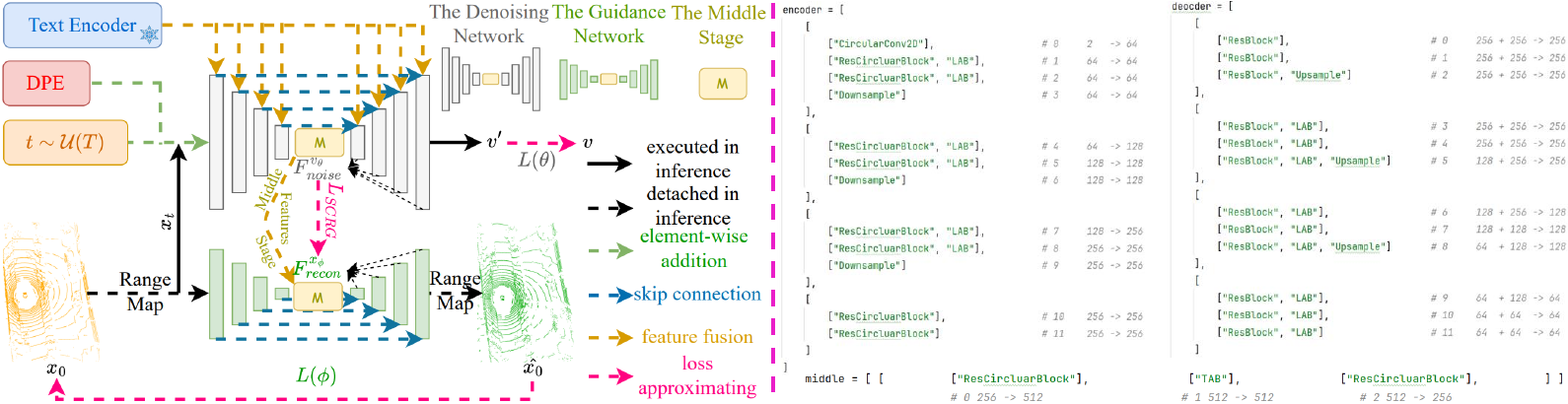}
	\caption{The overall framework of T2LDM++.  Left: The frozen Text Encoder (TE) encodes the text prompt, producing  semantically reliable features. Meanwhile, the auxiliary Guidance Network (GN) aligns with the real coordinates, providing the feature-level soft supervision. Furthermore, the dominant Denoising Network (DN) models the denoising process under text guidance, DPE, timestep and geometry-aware regularization, dominating the generation results. Right: \textit{\textbf{ResCircularBlock}} and \textit{\textbf{Resblock}} follow the standard residual block design. Meanwhile, we achieve \textit{\textbf{Downsampling}} by CircularConv2D (stride=2). Furthermore, the \textit{\textbf{Upsampling}} module consists of a bilinear interpolation layer and a Conv2D layer.}
	\label{fig6}
	
\end{figure*}

\subsection{Self-Conditioned Representation Guidance} 
\label{sec34}

Due to the high cost of collection process \cite{caccia2019deep, zyrianov2022learning, nakashima2024lidar, wu2024text2lidar, qu2026self}, high-quality LiDAR data remains significantly scarce. This often leads to insufficient training priors for the generative model, resulting in the lack of realistic, diverse, and detailed objects in generated LiDAR scenes (see Fig.~\ref{fig1} and Fig.~\ref{fig2}). Along another research line, image generation methods leverage representation priors to improve generation performance (see Sec.~\ref{sec23}), achieving significant results \cite{li2023self, yu2024representation}. Despite being effective, these methods typically require: 

\begin{itemize}
	\item Large-scale pretrained knowledge priors \cite{oquab2023dinov2}.
	\item Multi-stage training costs \cite{li2023self}.
	
\end{itemize}

In this paper, we propose a Self-Conditioned Representation Guidance (SCRG). SCRG employs a  Guidance Network (GN, $x_{\phi}$), detached in inference, to guide the Denoising Network (DN, $v_{\theta}$) in learning geometry-aware representations in an End-to-End manner. By the geometric regularization, DN can accurately reconstruct noise targets, approximating a reconstruction-consistent trajectory in inference (see Sec.~\ref{sec41}).

Specifically,  GN perceives multi-level perturbation features $F^{v_{\theta}}_{noise}$ from DN, reconstructing scene details by aligning the real coordinate ($\bm{x_0}$):

\begin{equation}
	\begin{split}
		\label{f341}
		L(\phi) =||\bm{x_0} - x_\phi(\bm{x_0},F^{v_{\theta}}_{noise})||^2.
	\end{split}
\end{equation}

Subsequently, $F^{v_{\theta}}_{noise}$ is aligned with the reconstruction signals $F^{x_{\phi}}_{recon}$ provided by GN, injecting geometry-aware regularization into DN: 

\begin{equation}
	\begin{split}
		\label{f342}
		L_{SCRG} =l_{recon}(F^{x_{\phi}}_{recon} - F^{v_{\theta}}_{noise}), 
	\end{split}
\end{equation}
where $l_{recon}(\cdot)$ is the cosine distance. Meanwhile, as mentioned in Sec.~\ref{sec32}, $L_{SCRG}$ should constrain DN at Decoder, further enhancing the high-frequency reconstruction process.

This simple and effective approach:
\begin{itemize}
	\item \textbf{Without Additional Training Priors.} GN is trained in an End-to-End manner, avoiding additional training samples and multi-stage training cost.
	\item \textbf{Without Inference Cost.} The detachable design of GN prevents cost and information leakage in inference.
	\item \textbf{With Faster Convergence.} The regularization from GN guides DN to learn high-frequency details for faster early-stage convergence using less training iterations.
\end{itemize}


\subsection{Network Architecture} 
\label{sec35}

In this section, we present the overall framework of T2LDM++, as illustrated in Fig.~\ref{fig6}. T2LDM++ consists of three key components: the frozen Text Encoder (TE), the lightweight Guidance Network (GN), and the dominant Denoising Network (DN).

\textbf{The Frozen Text Encoder.} TE encodes the text prompt to provide a semantically meaningful conditional embedding. Similar to the previous version \cite{qu2026self}, we still use CLIP \cite{radford2021learning} (768-dim) as the Text Encoder, since this Text-Image alignment pretraining provides stronger vision perception-aware representations than T5 \cite{raffel2020exploring}.


\textbf{The Auxiliary Guidance Network.} GN provides geometry-aware regularization for DN. Unlike the dominant DN, GN should exhibit lightweight due to the guidance function (see Sec.~\ref{sec42}), consisting only of Conv2D layers, SiLU activations, and Downsampling/Upsampling layers. To provide perturbation-adaptive soft supervision, the architecture of GN is aligned with that of DN in Encoder and Decoder, adopting a four-stage U-Net (see Fig.~\ref{fig9}). Meanwhile, GN takes only RM as input, allowing GN to focus purely on learning reconstruction details.

\begin{figure*}[htp]
	\centering
	\includegraphics[width=0.99\textwidth]{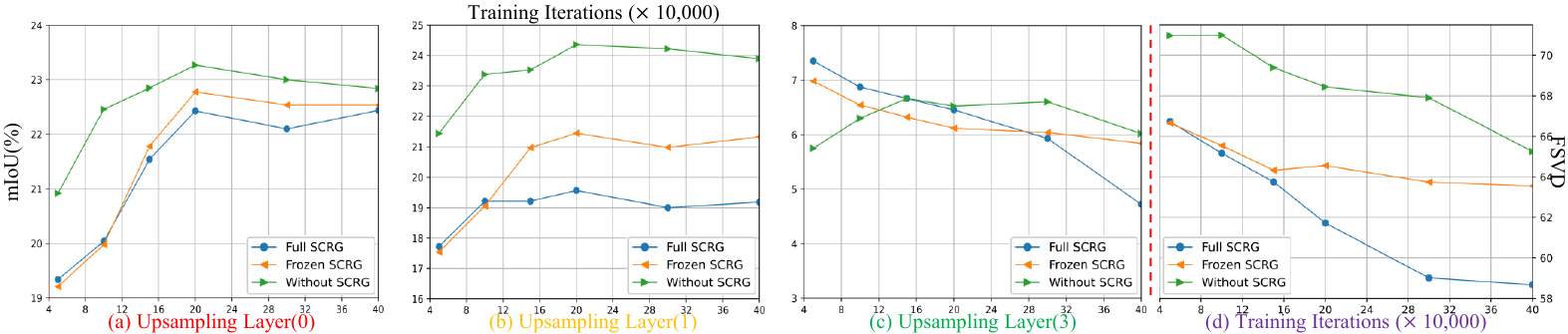}
	\caption{ The results based on enhanced DN (see Sec.~\ref{sec32}). (a) Upsampling Layer(0) means that the middle-layer output of DN is used as the prior features for SLP. (b) Upsampling Layer(1) means that the output of the first upsampling layer in DN is used as the prior features for SLP. (c) Upsampling Layer(3) means that the output of the last layer in Decoder of DN is used as the prior features for SLP. (d) Generation performance under different training iterations.}
	\label{fig7}
	
\end{figure*}

\begin{figure*}[htp]
	\centering
	\includegraphics[width=0.99\textwidth]{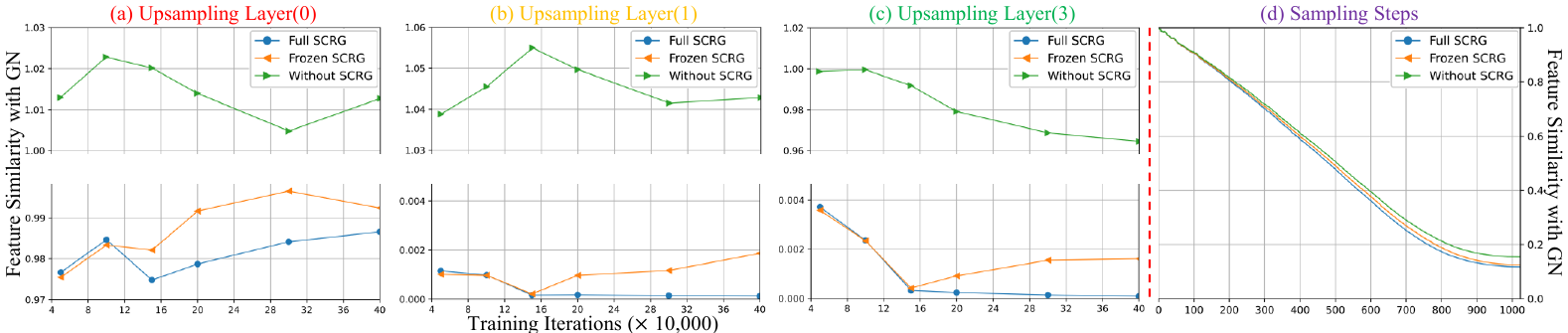}
	\caption{ The results based on enhanced DN (see Sec.~\ref{sec32}). (a) Upsampling Layer(0) means the similarity between the middle-layer output of DN and the corresponding features of GN. (b) Upsampling Layer(1) means the similarity between the output of the first upsampling layer in DN and the corresponding features of GN. (c) Upsampling Layer(3) means the similarity between the output of the last layer in the decoder of DN and the corresponding features of GN. (d) Under the same randomness, the feature similarity between DN and the corresponding GN along the sampling trajectory.}
	\label{fig8}
	
\end{figure*}

\textbf{The Dominant Denoising Network.} DN models the denoising process, dominating the generation results. This follows the U-Net architecture \cite{rombach2022high} (see Sec.\ref{sec31}), comprising four stages with Attention Block (AB) and Residual Block (RB) in Encoder and Decoder.

Specifically, AB takes the text features $F_{text}$ from TE to provide conditional guidance. Then, a cross-attention block fuses the projected features $F^{v_{\theta}}_{noise} \in \mathbb{R}^{l \times C^{v_{\theta}}}$$\rightarrow$$(Q) \in \mathbb{R}^{l \times C}$ and $F_{text} \in \mathbb{R}^{n \times 768}$$\rightarrow$$(K,V) \in \mathbb{R}^{n \times C}$:

\begin{equation}
	\begin{split}
		\label{f351}
		O = mlp(WV)+F^{v_{\theta}}_{noise}, \\
		F = ffn(O)+O, \quad\quad
	\end{split}
\end{equation}
where $W \in \mathbb{R}^{l \times n} = softmax(\frac{QK^T}{\sqrt{C}})$ is an attention weight map and $l=h \times w$.

Meanwhile, \textit{$F_{text}$=$F^{v_{\theta}}_{noise}$  means unconditional generation (self-attention).} The timestep $t$ and DPE are incorporated into RB to indicate the denoising stage and improve fidelity.

\subsection{Training and Inference} 
\label{sec36}

\textbf{Training.} As mentioned earlier, T2LDM++ models the denoising process within DN, while learning the geometry-aware representation via SCRG. Therefore, the training objective is:

\begin{equation}
	\begin{split}
		\label{f361}
		L_{total}=L(\theta)+L(\phi)+\lambda L_{SCRG},
	\end{split}
\end{equation}
 where $\lambda$ represents an epoch-wise weight \cite{qu2026self}. 

\textbf{Inference.} Benefiting from the detached design, T2LDM++ iteratively transforms $\bm{x_T}$ into $\bm{x'_0}$ using only $v_\theta$, independently of $x_\phi$:

\begin{equation}
	\begin{split}
    	\label{f362}
		\bm{x_{t-1}} = \frac{1}{\sqrt{\alpha_t}} (\bm{x_t} - \frac{1-\alpha_t}{\sigma_t} [ \sqrt{1-\bar{\alpha}_t} \bm{x_t} \quad\;
        \\
        +\sqrt{\bar{\alpha}_t} \, v_\theta] ) + \sqrt{\frac{1 - \bar{\alpha}_{t-1}}{1 - \bar{\alpha}_t}(1-\alpha_t)} \bm{\epsilon},
	\end{split}
\end{equation}

\section{Analysis and Optimization for SCRG} 
\label{sec4}

In this section, we analyze the underlying reasons that SCRG effectively improves generation performance. Furthermore, based on these insights, we redesign a more efficient and effective GN.

\begin{table*}[h]
	\resizebox{0.99\textwidth}{!}{
			\begin{tabular}{p{3.5cm}p{3.5cm}p{3.5cm}p{3.5cm}p{3.5cm}}	
				\Xhline{1pt}
				
				{Methods}
				&\makecell[c]{Full SCRG}
				&\makecell[c]{Frozen SCRG \cite{qu2026self}}
                &\makecell[c]{Without SCRG}
                &\makecell[c]{Without Pre.}\\

                \hline
                
				{mIoU($\%$)}
				&\makecell[c]{22.44}
				&\makecell[c]{22.78}
                &\makecell[c]{24.36}
                &\makecell[c]{1.3}\\
                
				\Xhline{1pt}
				
			\end{tabular}
		}
		\caption{The best segmentation results with different pretrained models on nuScenes for SLP. 'Frozen SCRG' means that GN is trained for the first 100K iterations and then frozen \cite{qu2026self}. Meanwhile, 'Without Pre.' means training from scratch without pretrained models. Using pretrained T2LDM++ significantly improves the segmentation results of SLP. 1) DDPM pretraining enhances representations. 2) SCRG improves feature discriminability (reconstruction).}
	\label{tab411}
\end{table*}

\subsection{How Does SCRG improve the performance in DDPMs?}
\label{sec41}

\textbf{Feature Discriminability in DN.} Unlike Classification Linear Probes \cite{xiang2023denoising} evaluating global representations, we adopt a Segmentation Linear Probe (SLP) to investigate point-wise feature discriminability across different models. Specifically, we extract features from different layers (Upsampling Layer(0) (UL0), Upsampling Layer(1) (UL1), and Upsampling Layer(3) (UL3)) of a pretrained DN as prior features. Unlike \cite{xiang2023denoising, wei2023diffusion, zheng2024point,chen2024deconstructing}, we adapt \textit{a more fundamental formulation}. Specifically, this adds noise to $\bm{x_0}$ with a minimal timestep ($t=0.0001$, $t\in[0.0001,0.02]$) as the input to the pretrained DN in training, while the clean $\bm{x_0}$ is used in inference. Meanwhile, for the segmentation head, we only use a one-layer MLP (the bilinear interpolation is applied when the insufficient resolution) to evaluate segmentation performance on nuScenes.  Tab.~\ref{tab411} shows that SLP with SCRG achieves lower segmentation results than that without SCRG. In fact, semantic segmentation inherently requires local semantic consistency. Features with strong discriminability (rich reconstruction details) often lead to poor performance in semantic tasks (e.g., classification, segmentation and detection) \cite{hu2011bridging, sun2025pixel}, as they focus on point-wise high-frequency details, losing local semantic consistency. For example, in a scene, features of a class should be distinguishable from those of other classes, while maintaining similarity within the class. If features within a class exhibit strong discriminability, this will blur intra-class homogeneity, leading to degraded segmentation performance. Therefore, \textit{SCRG enables DN to accurately predict the denoising targets by enhancing the reconstruction ability, due to using a reconstruction loss in DDPMs}. Essentially, \textit{denoising learning in DDPMs is a reconstruction task}. Fig.~\ref{fig7} further validates this analysis. Although Full SCRG exhibits lower segmentation performance than Without SCRG (see Fig.~\ref{fig7}(a)(b)(c)), Full SCRG achieves better generation performance (see Fig.~\ref{fig7}(d)). This indicates that SCRG enhances the feature discriminability of DN, leading to better reconstruction target prediction and distribution matching.

\textbf{Guidance in Training and Inference.} SCRG leverages GN to guide DN to learn geometry-aware representations during the denoising process in training and inference, improving generation performance. In fact, this guidance should keep a co-adaptation process between GN and DN, where the reconstruction supervision evolves dynamically with the representations of DN. Freezing GN in training may cause the supervision to become mismatched with the continuously updated DN representations, weakening the guidance effect, limiting model performance. Fig.~\ref{fig8} further supports this conclusion. Full SCRG consistently exhibits higher feature similarity to GN throughout training (see Fig.~\ref{fig8}(a)(b)(c)). This enables T2LDM++ to sample along the trajectory of the reconstruction-consistent representation (the $\bm{x_0}$ features captured by GN, see Fig.~\ref{fig8}(d)) in inference, thereby generating objects with richer details in scenes.

\begin{table*}[h]
  \resizebox{1.0\textwidth}{!}{
        \begin{tabular}{p{3.1cm}|p{2.2cm}p{2.2cm}p{2.2cm}p{2.0cm}p{2.2cm}p{2.2cm}}	
        \Xhline{1pt}
  
        {Methods}
        &\makecell[c]{$\#$Tra.Par.$\downarrow$}
        &{\makecell[c]{Tra.Tim.$\downarrow$}}
        &{\makecell[c]{Tra.Mem.$\downarrow$}}
        &{\makecell[c]{$\#$Inf.Par.$\downarrow$}}
        &{\makecell[c]{FSVD$\downarrow$}}
        &{\makecell[c]{FPVD$\downarrow$}} \\
        
       \hline

        {Without SCRG}
        &\makecell[c]{30.6M}
        &{\makecell[c]{$\sim$41h}}
        &{\makecell[c]{$\sim$3.5G}}
        &{\makecell[c]{30.6M}}
        &{\makecell[c]{65.25}}
        &{\makecell[c]{67.46}} \\

        {Frozen SCRG \cite{qu2026self}}
        &\makecell[c]{59.3M/30.6M}        &{\makecell[c]{$\sim$47h}}
        &{\makecell[c]{$\sim$5.9G/3.5G}}
        &{\makecell[c]{30.6M}}
        &{\makecell[c]{63.55}}
        &{\makecell[c]{65.71}} \\

        {Full SCRG}
        &\makecell[c]{59.3M}
        &{\makecell[c]{$\sim$63h}}
        &{\makecell[c]{$\sim$6.6G}}
        &{\makecell[c]{30.6M}}
        &{\makecell[c]{60.47}}
        &{\makecell[c]{63.32}} \\

        {Redesigned SCRG}
        &\makecell[c]{41.0M}
        &{\makecell[c]{$\sim$45h}}
        &{\makecell[c]{$\sim$4.3G}}
        &{\makecell[c]{30.6M}}
        &{\makecell[c]{58.68}}
        &{\makecell[c]{61.09}} \\

        \Xhline{1pt}
        
	\end{tabular}
 }
	\caption{Comparison of SCRG with different training schemes and architectures. The carefully redesigned GN achieves lower cost and better generation performance. All models are trained on nuScenes using 8 NVIDIA RTX 4090 GPUs.}
	\label{tab422}
\end{table*}

\textbf{Results and Insights.} Based on the above analysis, we obtain the following insights from Tab.~\ref{tab411}, Fig.~\ref{fig7} and Fig.~\ref{fig8}:

\begin{itemize}
	\item With pretrained DN, the segmentation performance of SLP is significantly improved (1.3$\%$ $\rightarrow$ 24.36$\%$, see Tab.~\ref{tab411}). \textit{\textbf{This indicates that DDPMs, as a pretrained paradigm in a return-to-essence manner, have great potential to be explored.}}
    
	\item As the decoder depth increases (increased feature discriminability), the segmentation performance of SLP significantly decreases (see Fig.~\ref{fig7}(a)(b)(c)). This verifies the nature of features: \textit{\textbf{as reconstruction ability improves, semantic consistency decreases \cite{hu2011bridging, sun2025pixel}}}.
    
	\item The best segmentation performance is achieved at UL1 of DN,  consistent with the conclusion of \cite{wei2023diffusion} (see Fig.~\ref{fig7}(b)). However, our pretrained strategy is more consistent with the mathematical formulation, since \textit{\textbf{the input corresponds to 'the truly clean data' for DDPMs in the stage of training SLP ($t$=0.0001 indicates the minimal time step)}}. 
    
    \item With the application of SCRG, SLP performance decreases, while generation performance significantly improves (see Fig.~\ref{fig7}(a)(b)(c)(d) and Fig.~\ref{fig8}). This indicates that SCRG essentially improves generation performance by \textit{\textbf{enhancing the reconstruction ability (detailed representations) of the generative model.}} This is crucial for accurate denoising, due to the reconstruction loss in DDPMs. 
    
\end{itemize}

Intuitively, the training objective of DDPMs is essentially a form of “reconstruction task”, due to employing a reconstruction loss. More fundamentally, the model matches the real data distribution through reconstructing the score ($\bm{\epsilon}$, $\bm{x_0}$, and $\bm{v}$) \cite{qu2025end}. Therefore, improving reconstruction ability of the denoising network leads to more accurate denoising, yielding better generation performance.

\subsection{Optimization of SCRG}
\label{sec42}

In T2LDM \cite{qu2026self}, GN adopts the same network architecture with DN. In practice, based on the above analysis, we can further optimize the design of GN.

\textbf{Full vs. Frozen.} As mentioned in Sec~\ref{sec41} and shown in Fig.~\ref{fig8}, GN should be jointly trained with DN throughout the entire process, so that the guidance remains continuously aligned with the representations of DN. 

\begin{figure}[htp]
	\centering
	\includegraphics[width=0.48\textwidth]{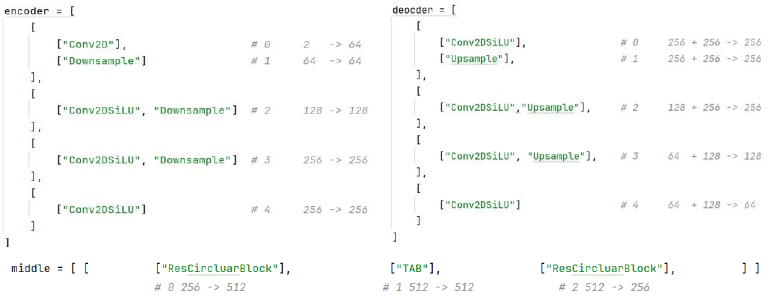}
	\caption{The network architecture of GN. This mainly consists of the \textit{\textbf{Downsampling}} module (a Conv2D layer with stride$=$2), the \textit{\textbf{Upsampling}} modules (a bilinear interpolation layer), and the \textit{\textbf{Conv2DSiLU}} layer (a Conv2D layer with a SiLU activation function). To align with Encoder and Decoder in DN, GN also adopts a 4-stage U-Net.}
	\label{fig9}
	
\end{figure}

\textbf{Noise Feature Interaction Position.} To ensure that GN purely aligns with real representations while adapting to multi-level perturbations from DN, we follow the design of feature interaction at the bottleneck layer with the largest receptive field \cite{huang2021predator, qu2024conditional, qu2025end}. This minimizes the disturbance of noise features to reconstruction details in GN, focusing purely on learning clean reconstruction representations.

\textbf{Network Capacity.} Similarly, as discussed in Sec.~\ref{sec41} and shown in Fig.~\ref{fig8}, GN provides the geometry-aware regularization, guiding the denoising process to evolve along reconstruction-consistent directions, improving generation quality. Therefore, a lightweight GN should  provide enough guidance, as this mainly lies in the constraint form of SCRG. In fact, since DN plays the dominant role in the task, GN should be lightweight. A strong representational capacity of GN may dominate the training process, weakening the representation learning of DN \cite{qu2025end}.

Based on the above, we carefully refine the design of GN:

\begin{itemize}
	\item GN should adopt a full training scheme to adapt to the representations of DN.
    
	\item Only the attention in the middle stage is retained to perceive multi-level noise features from DN.
    
	\item ResBlock is simplified into a combination of a MLP layer and a SiLU activation function ($\sim$10M total parameters). 

\end{itemize}

The overall architecture of GN is shown in Fig.~\ref{fig9}. Meanwhile, Tab.~\ref{tab422} presents a comparison of different training schemes and network architectures of GN.

\section{Text-LiDAR Benchmark}
\label{sec5}

\begin{table*}[h]
  \resizebox{1.0\textwidth}{!}{
        \begin{tabular}{p{2.1cm}|p{2.2cm}|p{1.2cm}p{1.2cm}p{1.0cm}p{1.2cm}p{1.2cm}p{1.2cm}p{1.2cm}p{1.0cm}p{1.0cm}p{1.0cm}}	
        \Xhline{1pt}
  
        {Benchmark}
        &\makecell[c]{Sample/Text}
        &{\makecell[c]{barrier}}
        &{\makecell[c]{bicycle}}
        &{\makecell[c]{bus}}
        &{\makecell[c]{car}}
        &{\makecell[c]{vehicle}}
        &{\makecell[c]{motor}}
        &{\makecell[c]{pedes.}}
        &{\makecell[c]{cone}}
        &{\makecell[c]{trailer}}
        &{\makecell[c]{truck}} \\
        
       \hline

        nuScenes\cite{caesar2020nuscenes}
        &\makecell[c]{34149/-}
        &\makecell[c]{12320}
        &\makecell[c]{7474}
        &\makecell[c]{10986}
        &\makecell[c]{33266}
        &\makecell[c]{9495}
        &\makecell[c]{7518}
        &\makecell[c]{27862}
        &\makecell[c]{14853}
        &\makecell[c]{9432}
        &\makecell[c]{24118}\\

        T2nuScenes\cite{qu2026self}
        &\makecell[c]{34149/8}
        &\makecell[c]{3819}
        &\makecell[c]{-}
        &\makecell[c]{-}
        &\makecell[c]{21876}
        &\makecell[c]{-}
        &\makecell[c]{-}
        &\makecell[c]{11534}
        &\makecell[c]{-}
        &\makecell[c]{-}
        &\makecell[c]{6523}\\

        T2nuScenes++
        &\makecell[c]{150883/65}
        &\makecell[c]{10709}
        &\makecell[c]{7071}
        &\makecell[c]{10533}
        &\makecell[c]{131883}
        &\makecell[c]{7861}
        &\makecell[c]{7266}
        &\makecell[c]{26397}
        &\makecell[c]{13294}
        &\makecell[c]{8380}
        &\makecell[c]{23563}\\

        \Xhline{1pt}
        
	\end{tabular}
 }
	\caption{The class distribution of nuScenes \cite{caesar2020nuscenes},  T2nuScenes \cite{qu2026self} and T2nuScenes++. 'Sample/Text' denotes the number of samples/text types. We categorize classes into foreground classes (barrier, bicycle, bus, car, vehicle, motor, pedestrian, cone, trailer, truck) and background classes (surface, flat, sidewalk, terrain, manmade, vegetation).  T2nuScenes++ closely follows the real-world distribution (nuScenes). All text types are provided in the supplementary material \ref{app_sec11}.}
	\label{tab511}
\end{table*}

\begin{table*}[h]
  \resizebox{1.0\textwidth}{!}{
        \begin{tabular}{p{3.0cm}|p{2.2cm}|p{2.2cm}p{2.2cm}p{2.0cm}|p{1.2cm}p{1.2cm}p{1.2cm}|p{1.2cm}p{1.0cm}p{1.0cm}p{1.0cm}}	
        \Xhline{1pt}

        \multirow{2}{*}{Benchmark}
        &\multirow{2}{*}{Sample/Text}
        &\multicolumn{3}{c|}{Low-Frequency Class}
        &\multicolumn{3}{c|}{Mid-Frequency Class}
        &\multicolumn{4}{c}{High-Frequency Class}\\

        &
        &{\makecell[c]{motorcyclist}}
        &{\makecell[c]{bicyclist}}
        &{\makecell[c]{motorcycle}}
        &{\makecell[c]{vehicle}}
        &{\makecell[c]{bicycle}}
        &{\makecell[c]{person}}
        &{\makecell[c]{pole}}
        &{\makecell[c]{sign}}
        &{\makecell[c]{trunk}}
        &{\makecell[c]{car}} \\
        
       \hline

        SemanticKITTI\cite{behley2019semantickitti}
        &\makecell[c]{34149/-}
        &\makecell[c]{719}
        &\makecell[c]{2230}
        &\makecell[c]{3820}
        &\makecell[c]{7322}
        &\makecell[c]{5379}
        &\makecell[c]{6851}
        &\makecell[c]{22688}
        &\makecell[c]{16112}
        &\makecell[c]{21062}
        &\makecell[c]{21855}\\

        T2SemanticKITTI
        &\makecell[c]{127140/50}
        &\makecell[c]{2507}
        &\makecell[c]{3315}
        &\makecell[c]{3748}
        &\makecell[c]{8271}
        &\makecell[c]{7965}
        &\makecell[c]{8235}
        &\makecell[c]{11936}
        &\makecell[c]{11966}
        &\makecell[c]{11219}
        &\makecell[c]{29767}\\

        \Xhline{1pt}
        
	\end{tabular}
 }
	\caption{The class distribution of SemanticKITTI and T2SemanticKITTI. We categorize classes into foreground classes (motorcyclist, bicyclist, motorcycle, vehicle, bicycle, person, pole, sign, trunk, car) and background classes (road, parking, sidewalk, building, fence, vegetation, terrain, ground). All text types are provided in the supplementary material \ref{app_sec12}.}
	\label{tab521}
\end{table*}

In this section, we focus on constructing Text-LiDAR benchmarks, hoping to provide effective insights in Text-to-LiDAR generation tasks. Based on the existing approach \cite{qu2026self}, we construct T2nuScenes++ (150,883 samples) using 3D Boxes and T2SemanticKITTI (127,140 samples) using Semantic Labels, respectively.

\subsection{T2nuScenes++} 
\label{sec51}

\begin{figure}[htp]
	\centering
	\includegraphics[width=0.48\textwidth]{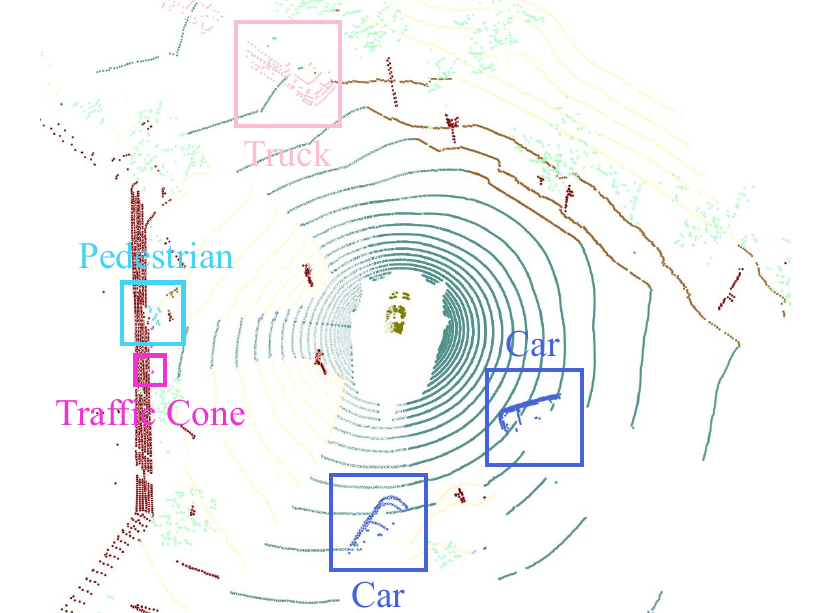}
	\caption{Using the geometric-annotation-based strategy, a single LiDAR scene can correspond to multiple types of text descriptions: (a) Two cars. (b) One car is to the right of one pedestrian. (c) The scene contains a car and a truck. (d) Cars and traffic cones. (e) One car is facing right. }
	\label{fig10}
	
\end{figure}

Following T2nuScenes \cite{qu2026self}, T2nuScenes++ also focuses on “car” as the target class, as this represents the primary object and the stable structure in LiDAR scenes. Inspired by the geometric annotation construction strategy \cite{qu2026self}, a single scene can actually correspond to multiple types of text descriptions. Therefore, this allows us to produce multiple Text-LiDAR pairs for a LiDAR scene (see Fig.~\ref{fig10}), further expanding the benchmark.


Meanwhile, to improve the quality of scene descriptions, we further introduce a series of refinements for T2nuScenes++: 1) We categorize classes into \textit{foreground objects (retained) and background elements (removed)}, to reduce semantic ambiguity. 2) To better reflect the real-world distribution, we adjust \textit{the class distribution according to the occurrence ratios of each class in nuScenes}. 3) We retain \textit{diverse description types for the target class (position, quantity, orientation, and relationships with other classes)}, to enhance text generalization. 

Based on the above, we construct a large-scale, real-world distribution-aligned and description-diverse Text-LiDAR benchmark, T2nuScenes++, containing 150,883 samples. The class distribution comparison of T2nuScenes \cite{qu2026self} and T2nuScenes++ is shown in Tab.~\ref{tab511}.

\begin{figure*}[htp]
	\centering
	\includegraphics[width=1.0\textwidth]{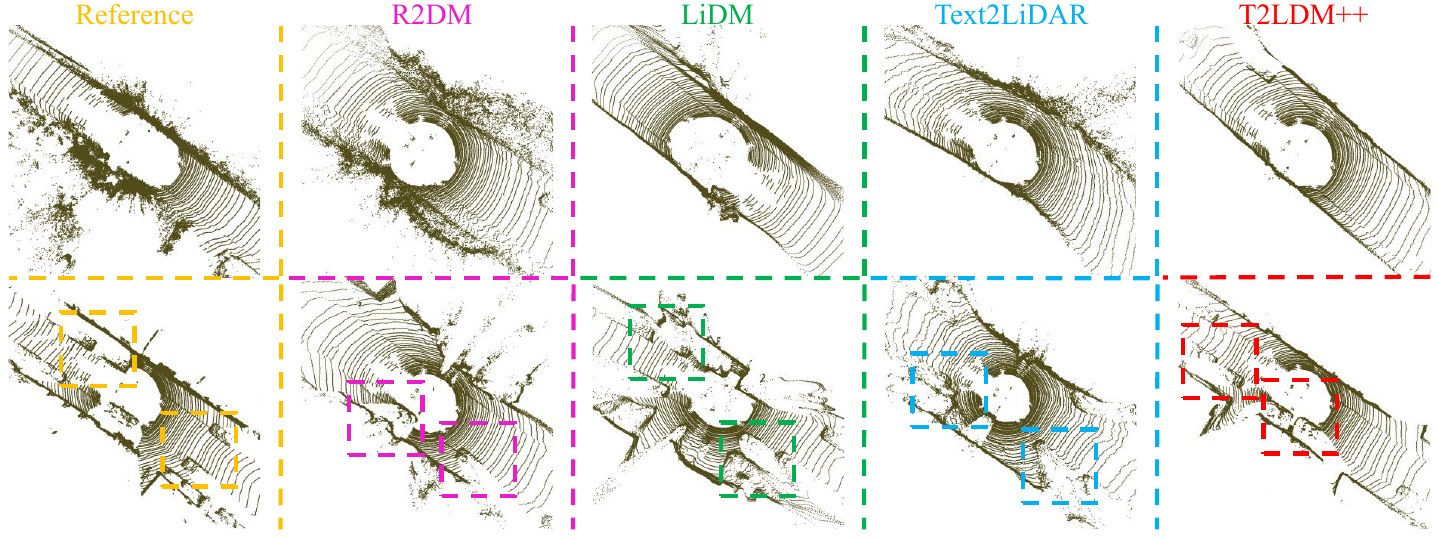}
	\caption{The unconditional generation results on KITTI360. Most methods achieve promising results in simple scenes (first row), but struggle in complex scenes. In contrast, T2LDM++ generates results closer to the real-world distribution with clearer geometric details even in  multi-object scenes (second row).}
	\label{fig11}
	
\end{figure*}

\subsection{T2SemanticKITTI}
\label{sec52}

Furthermore, we extend the text description from using 3D Boxes to utilizing Semantic Labels. However, unlike 3D Boxes with diverse geometric priors, Semantic Labels lack explicit encoding of the locations of scene objects, struggling to capture the position relationships between objects. Therefore, we construct text descriptions uniformly for all classes (but focusing on the target class 'car'). Meanwhile, we enrich the diversity of descriptions through class combinations. 

Meanwhile, to balance the class distribution, similar to T2nuScenes++, we first remove background categories in SemanticKITTI. Then, we categorize foreground classes into three groups based on the sample occurrence frequency: low-frequency ($<$4,000), mid-frequency (4,000-10,000), and high-frequency ($>$10,000) (see Tab.~\ref{tab521}). We augment text for low-frequency classes, retain descriptions for mid-frequency classes, and downsample for high-frequency classes to ensure a balanced class distribution. This alleviates the long-tail problem in the original class distribution.

Based on this, we construct a Text-LiDAR benchmark using Semantic Labels, T2SemanticKITTI, comprising 127,140 samples. Although Semantic Labels provide less geometric priors than 3D Boxes, this strategy still enables constructing meaningful scene descriptions, demonstrating the generalizability.

Overall, geometric annotations are deterministic and physically grounded, essentially serving as a form of geometric semantics. This reduces the ambiguity introduced by free-form human descriptions. We hope this can inspire future research on leveraging geometric annotations as a unified source for constructing controllable Text-LiDAR benchmarks and text-guided 3D scene generation.

\section{Experiments}\label{sec6}

All codes, demos and results of T2LDM++ are available at this \href{https://github.com/QWTforGithub/T2LDM_v2}{website}.

\subsection{Experiment Setup}

\textbf{Dataset.} We conduct evaluations on three public datasets: KITTI-360 \cite{liao2022kitti}, SemanticKITTI \cite{behley2019semantickitti}, and nuScenes \cite{caesar2020nuscenes}. KITTI-360 and SemanticKITTI are collected using 64-beam LiDARs with $360^\circ$ surround-view scanning, containing 76,165 and 23,021 samples for LiDAR scene generation. Similarly, nuScenes is built using a 32-beam LiDAR to capture full-surround scenes, including 34,149 samples for training. 

\textbf{Metric.} Following the previous version \cite{qu2026self}, we adopt FID (FSVD, FPVD), JSD, and MMD as evaluation metrics. Meanwhile, for fair
comparison, we compute the real distribution mean and variance of FID using all samples from KITTI-360, SemanticKITTI, and nuScenes. Furthermore, we propose a controllability metric for text-guided generation, the matching \textbf{R}ate between the \textbf{T}ext prompt and the predicted 3D \textbf{B}oxes (TBK). This employs a detector \cite{liu2025fshnet} to predict 3D boxes from generated results, measuring the matching rate between text semantics and generated scenes:

\begin{equation}
	\begin{split}
    	\label{f611}
		TBK=\frac{S_{match}}{S_{gen}},
	\end{split}
\end{equation}
where $S_{match}$ denotes the number of samples predicted 3D Boxes matching with text semantics. Meanwhile,  $S_{gen}$ denotes the number of total generated samples.

\subsection{Unconditional Generation}
\label{sec62}

\textbf{KITTI-360.} We first conduct evaluations on KITTI-360 captured using a 64-beam LiDAR. Since KITTI-360 contains more than 70,000 samples, we observe that evaluating FID using only 10,000 samples leads to unstable results. Therefore, we sample 30,000 samples for all models. As shown in Tab.~\ref{tab621}, T2LDM++ achieves lower FID, better matching the real distribution. This is because SCRG enhances the reconstruction representation through reconstruction-based soft supervision, enabling T2LDM++ to accurately reconstruct noise targets. Therefore, compared with existing methods with insufficient training priors, T2LDM++ can better perceive geometric structures from the real distribution in the sampling trajectory. Fig.~\ref{fig11} further provides qualitative results.

\begin{table}[h]
	\resizebox{0.475\textwidth}{!}{
			\begin{tabular}{p{2.5cm}|p{0.9cm}p{0.9cm}|p{1.0cm}p{1.1cm}p{1.0cm}p{1.0cm}}	
				\Xhline{1pt}
				
				{Methods}
				&\makecell[c]{GS}
				&\makecell[c]{RS}
				&\makecell[c]{FSVD$\downarrow$}
				&\makecell[c]{FPVD$\downarrow$}
				&\makecell[c]{JSD$\downarrow$}
				&\makecell[c]{MMD$\downarrow$}\\
				
				\Xhline{1pt}
				
				LiDARVAE \cite{caccia2019deep}
				&\makecell[c]{30000}
				&\makecell[c]{76165}      
				&\makecell[c]{285.76}
				&\makecell[c]{290.16}
				&\makecell[c]{0.36}
				&\makecell[c]{6.95}\\
				
				LiDARGAN \cite{caccia2019deep}
				&\makecell[c]{30000}
				&\makecell[c]{76165}       
				&\makecell[c]{345.55}
				&\makecell[c]{338.79}
				&\makecell[c]{0.38}
				&\makecell[c]{5.32}\\
				
				ProjectedGAN \cite{sauer2021projected}
				&\makecell[c]{30000}
				&\makecell[c]{76165}       
				&\makecell[c]{189.88}
				&\makecell[c]{203.75}
				&\makecell[c]{0.33}
				&\makecell[c]{3.46}\\
				
				LiDARGen \cite{zyrianov2022learning}
				&\makecell[c]{30000}
				&\makecell[c]{76165}       
				&\makecell[c]{240.14}
				&\makecell[c]{247.27}
				&\makecell[c]{0.33}
				&\makecell[c]{4.01}\\
				
				LiDM \cite{ran2024towards}
				&\makecell[c]{30000}
				&\makecell[c]{76165}       
				&\makecell[c]{205.44}
				&\makecell[c]{227.59}
				&\makecell[c]{0.34}
				&\makecell[c]{4.56}\\
				
				R2DM \cite{nakashima2024lidar}
				&\makecell[c]{30000}
				&\makecell[c]{76165}      
				&\makecell[c]{34.15}
				&\makecell[c]{38.55}
				&\makecell[c]{0.32}
				&\makecell[c]{4.17}\\
				
				Text2LiDAR \cite{wu2024text2lidar}
				&\makecell[c]{30000}
				&\makecell[c]{76165}      
				&\makecell[c]{55.14}
				&\makecell[c]{58.63}
				&\makecell[c]{0.33}
				&\makecell[c]{4.23}\\
                
				T2LDM \cite{qu2026self}
				&\makecell[c]{30000}
				&\makecell[c]{76165}        
				&\makecell[c]{25.74}
				&\makecell[c]{29.01}
				&\makecell[c]{0.30}
				&\makecell[c]{3.44}\\

				\cellcolor[rgb]{0.9725, 0.8078, 0.8} T2LDM++ 
				&\cellcolor[rgb]{0.9725, 0.8078, 0.8} \makecell[c]{30000}
				&\cellcolor[rgb]{0.9725, 0.8078, 0.8} \makecell[c]{76165}        
				&\cellcolor[rgb]{0.9725, 0.8078, 0.8} \makecell[c]{23.34}
				&\cellcolor[rgb]{0.9725, 0.8078, 0.8} \makecell[c]{26.86}
				&\cellcolor[rgb]{0.9725, 0.8078, 0.8} \makecell[c]{0.29}
				&\cellcolor[rgb]{0.9725, 0.8078, 0.8} \makecell[c]{3.23}\\
                
				\Xhline{1pt}
				
			\end{tabular}
		}

    \caption{The unconditional generation on KITTI-360. 'GS' means the generation samples. 'RS' denotes the real samples. Compared with other methods, the generation results of T2LDM++ are closer to the real distribution.}
    \label{tab621}
    \vspace{-20pt}
\end{table}

\begin{table}[h]
	\resizebox{0.475\textwidth}{!}{
			\begin{tabular}{p{2.5cm}|p{0.9cm}p{0.9cm}|p{1.0cm}p{1.1cm}p{1.0cm}p{1.0cm}}	
				\Xhline{1pt}
				
				{Methods}
				&\makecell[c]{GS}
				&\makecell[c]{RS}
				&\makecell[c]{FSVD$\downarrow$}
				&\makecell[c]{FPVD$\downarrow$}
				&\makecell[c]{JSD$\downarrow$}
				&\makecell[c]{MMD$\downarrow$}\\
				
				\Xhline{1pt}
								
				R2DM \cite{nakashima2024lidar}
				&\makecell[c]{10000}
				&\makecell[c]{23021}      
				&\makecell[c]{47.35}
				&\makecell[c]{49.62}
				&\makecell[c]{0.35}
				&\makecell[c]{4.77}\\
				
				Text2LiDAR \cite{wu2024text2lidar}
				&\makecell[c]{10000}
				&\makecell[c]{23021}      
				&\makecell[c]{65.67}
				&\makecell[c]{69.33}
				&\makecell[c]{0.36}
				&\makecell[c]{4.87}\\
                
				T2LDM \cite{qu2026self}
				&\makecell[c]{10000}
				&\makecell[c]{23021}        
				&\makecell[c]{27.88}
				&\makecell[c]{31.12}
				&\makecell[c]{0.33}
				&\makecell[c]{4.03}\\

				\cellcolor[rgb]{1.0, 0.9490, 0.8} T2LDM++ 
				&\cellcolor[rgb]{1.0, 0.9490, 0.8} \makecell[c]{10000}
				&\cellcolor[rgb]{1.0, 0.9490, 0.8} \makecell[c]{23021}        
				&\cellcolor[rgb]{1.0, 0.9490, 0.8} \makecell[c]{24.37}
				&\cellcolor[rgb]{1.0, 0.9490, 0.8} \makecell[c]{28.14}
				&\cellcolor[rgb]{1.0, 0.9490, 0.8} \makecell[c]{0.31}
				&\cellcolor[rgb]{1.0, 0.9490, 0.8} \makecell[c]{3.85}\\
                
				\Xhline{1pt}
				
			\end{tabular}
		}

    \caption{The unconditional generation on SemanticKITTI. T2LDM++ demonstrates superior generation performance with fewer training samples.}
    \label{tab622}
    \vspace{-20pt}
\end{table}

\begin{table}[h]
	\resizebox{0.475\textwidth}{!}{
			\begin{tabular}{p{2.5cm}|p{0.9cm}p{0.9cm}|p{1.0cm}p{1.1cm}p{1.0cm}p{1.0cm}}	
				\Xhline{1pt}
				
				{Methods}
				&\makecell[c]{GS}
				&\makecell[c]{RS}
				&\makecell[c]{FSVD$\downarrow$}
				&\makecell[c]{FPVD$\downarrow$}
				&\makecell[c]{JSD$\downarrow$}
				&\makecell[c]{MMD$\downarrow$}\\
				
				\Xhline{1pt}
				
				R2DM \cite{nakashima2024lidar}
				&\makecell[c]{10000}
				&\makecell[c]{34149}      
				&\makecell[c]{84.55}
				&\makecell[c]{89.29}
				&\makecell[c]{0.41}
				&\makecell[c]{4.98}\\
				
				Text2LiDAR \cite{wu2024text2lidar}
				&\makecell[c]{10000}
				&\makecell[c]{34149}      
				&\makecell[c]{81.37}
				&\makecell[c]{86.68}
				&\makecell[c]{0.34}
				&\makecell[c]{3.46}\\

                T2LDM \cite{qu2026self}
				&\makecell[c]{10000}
				&\makecell[c]{34149}        
				&\makecell[c]{61.98}
				&\makecell[c]{64.01}
				&\makecell[c]{0.26}
				&\makecell[c]{3.01}\\

				\cellcolor[rgb]{0.8352, 0.9098, 0.8314} T2LDM++ 
				&\cellcolor[rgb]{0.8352, 0.9098, 0.8314} \makecell[c]{10000}
				&\cellcolor[rgb]{0.8352, 0.9098, 0.8314} \makecell[c]{34149}        
				&\cellcolor[rgb]{0.8352, 0.9098, 0.8314} \makecell[c]{58.68}
				&\cellcolor[rgb]{0.8352, 0.9098, 0.8314} \makecell[c]{61.09}
				&\cellcolor[rgb]{0.8352, 0.9098, 0.8314} \makecell[c]{0.25}
				&\cellcolor[rgb]{0.8352, 0.9098, 0.8314} \makecell[c]{3.00}\\
                
				\Xhline{1pt}
				
			\end{tabular}
		}

    \caption{The unconditional generation on nuScenes.  T2LDM++ achieves state-of-the-art results  across all metrics in sparse scenes.}
    \label{tab623}
    \vspace{-20pt}
\end{table}

\begin{figure*}[htp]
	\centering
	\includegraphics[width=0.99\textwidth]{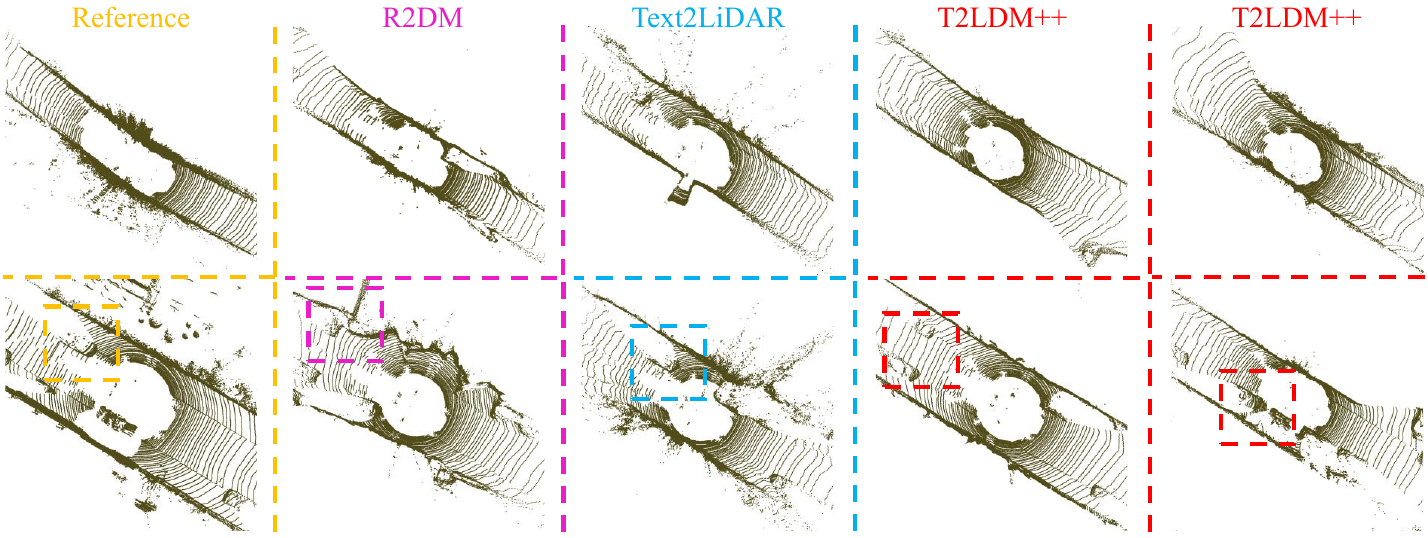}
	\caption{The unconditional generation results on SemanticKITTI. Insufficient training priors (limited samples) cause existing methods to generate over-smoothed results (second row). In comparison, benefiting from SCRG, T2LDM++ can generate detail-rich objects in complex scenes.}
	\label{fig12}
	
\end{figure*}

\textbf{SemanticKITTI.} Meanwhile, we conduct comparisons on SemanticKITTI with fewer samples. Tab.~\ref{tab622} shows that T2LDM++ still achieves superior generation performance compared with other methods. Benefiting from the geometric regularization of SCRG, T2LDM++ can effectively learn geometry-aware representations in the denoising learning, generating detailed scene objects under insufficient training priors. Meanwhile, Fig.~\ref{fig12} shows the superior generation results of T2LDM++.

\begin{figure*}[htp]
	\centering
	\includegraphics[width=0.99\textwidth]{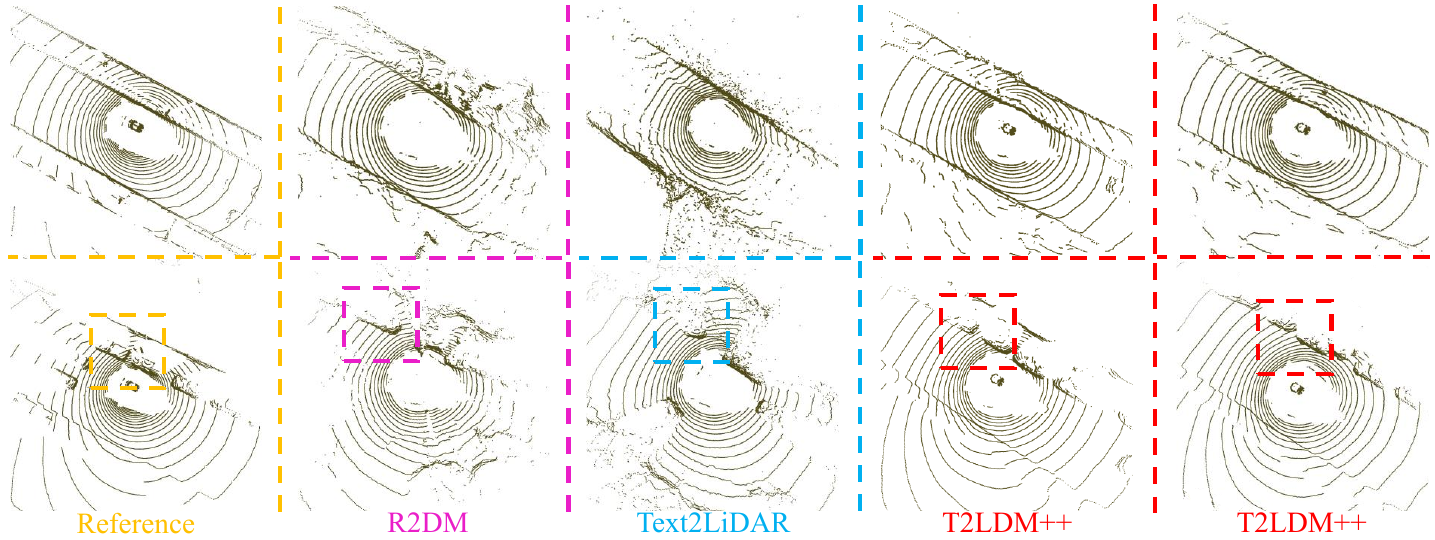}
	\caption{The unconditional generation results on nuScenes. In more challenging sparse scenes, most methods usually perform poorly (second row). However, guided by geometric regularization and directional priors, T2LDM++ can generate diverse (Fig.~\ref{fig2})  and realistic results.}
	\label{fig13}
	
\end{figure*}

\textbf{nuScenes.} Furthermore, we also conduct experiments on nuScenes with sparser point clouds. Compared to 64-Beam benchmarks, nuScenes with fewer points and a larger spatial distance between points is more challenging due to the harder-to-capture geometric details. This also causes existing methods to usually perform poorly on nuScenes. In contrast, T2LDM++ achieves remarkable generation results, as shown in Tab.~\ref{tab623}. Benefiting from the geometric priors of SCRG and DPE, T2LDM++ can effectively perceive geometric details from the real distribution. Fig.~\ref{fig2} and Fig.~\ref{fig13} further demonstrate that T2LDM++ can generate diverse and realistic LiDAR scenes.

\subsection{Text-Guided Generation} 
\label{sec63}

Compared with other conditions, text semantics are more directly obtained from humans, providing customized and diverse scene descriptions.

\begin{figure*}[htp]
	\centering
	\includegraphics[width=0.99\textwidth]{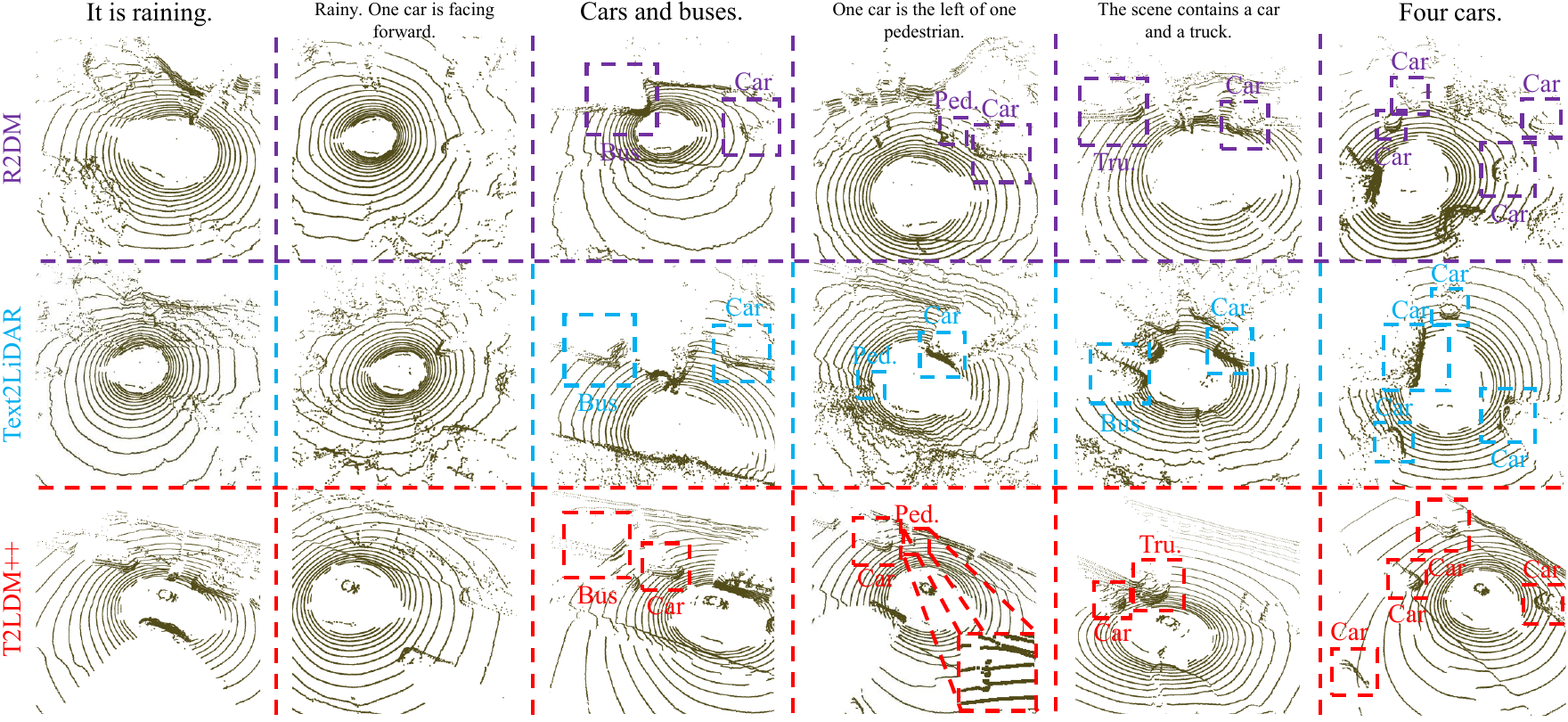}
	\caption{Text-guided generation results on T2nuScenes++. T2LDM++ shows  excellent controllability and  performance.}
	\label{fig14}
	
\end{figure*}

\begin{figure*}[htp]
	\centering
	\includegraphics[width=0.99\textwidth]{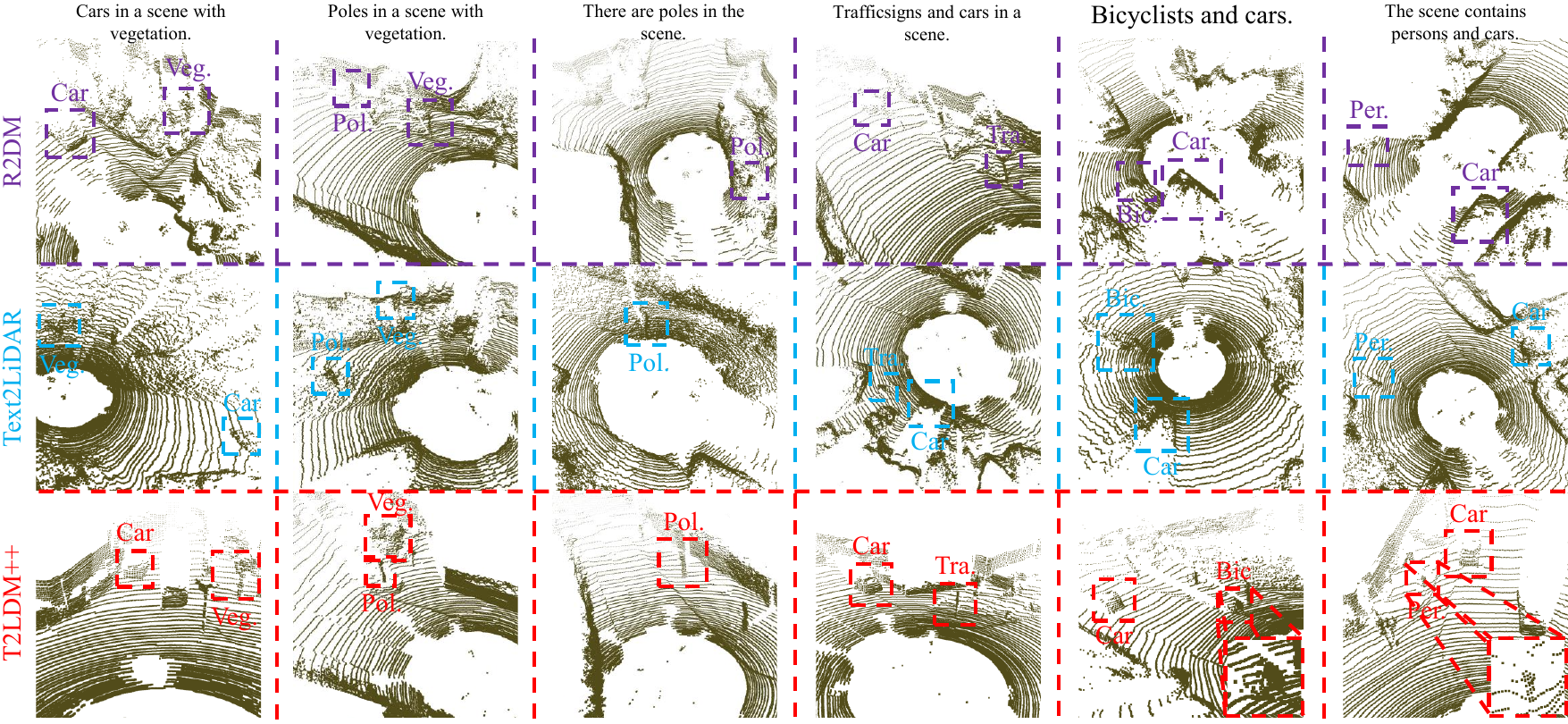}
	\caption{Text-guided generation results on T2SemanticKITTI. T2LDM++ can generate controllable LiDAR scenes from text semantics, producing objects with rich details.}
	\label{fig15}
	
\end{figure*}

\textbf{T2nuScenes++.} We conduct Text-to-LiDAR generation on T2nuScenes++. As mentioned in Sec.~\ref{sec51}, the text descriptions are derived from 3D Boxes. Therefore, we can measure the matching rate between text semantics and generated scenes through the detection prior, FSHNet \cite{liu2025fshnet}. Tab.~\ref{tab631} presents the results. T2LDM++ achieves the best controllability and generation quality. In fact, GN receives perturbed features with text conditions from DN, providing condition-guided feature supervision. Therefore, SCRG can enhance controllability in conditional generation, improving Text-to-LiDAR generation quality. Meanwhile, Fig.~\ref{fig14} shows qualitative Text-to-LiDAR generation results on T2nuScenes++. 

\textbf{T2SemanticKITTI.} Meanwhile, we also compare Text-to-LiDAR generation results on T2SemanticKITTI. We employ CDSegNet \cite{qu2025end} to evaluate the matching \textbf{R}ate between \textbf{T}ext prompt and the predicted \textbf{S}emantic label (TSR), due to the stronger noise robustness than existing segmentation models. Similarly, T2LDM++ still achieves superior results, as shown in Tab.~\ref{tab632}. GN participating throughout the full training process effectively improves controllability and generation quality. Furthermore, Fig.~\ref{fig15} illustrates the significant generation results of T2LDM++ on T2SemanticKITTI. This also demonstrates the generalization of constructing Text-to-LiDAR benchmarks based on geometric annotations.

\textbf{Zero-Shot Text-guided Generation.} Since the text descriptions are constructed from geometric annotations, they can be transformed into geometric representations for conditional generation. This means that the trained Box-to-LiDAR (see Sec.~\ref{sec64})  and Semantic-to-LiDAR (see Sec.~\ref{sec64}) T2LDM++ can directly achieve Zero-Shot Text-to-LiDAR generation. Meanwhile, we encode the input text prompt and all text descriptions in benchmarks using T5 \cite{raffel2020exploring}, retrieving the geometric representation with the most similar semantics via $L_2$ distance. Tab.~\ref{tab631} and Tab.~\ref{tab632} present the results of Zero-Shot Text-guided generation on T2nuScenes and T2SemanticKITTI, respectively. Converting text into geometric representations significantly improves generation quality and controllability (see Tab.~\ref{tab632}), as they provide fine-grained scene geometric structures, enabling more precise conditional guidance during generation. This inspires a new feasible pipeline, hoping to encourage further exploration of Text-to-LiDAR generation tasks.

\begin{table}[h]
	\resizebox{0.475\textwidth}{!}{
			\begin{tabular}{p{2.5cm}|p{0.9cm}p{0.9cm}|p{1.0cm}p{1.1cm}p{1.0cm}p{1.0cm}p{1.0cm}}	
				\Xhline{1pt}
				
				{Methods}
				&\makecell[c]{GS}
				&\makecell[c]{RS}
				&\makecell[c]{FSVD$\downarrow$}
				&\makecell[c]{FPVD$\downarrow$}
				&\makecell[c]{JSD$\downarrow$}
				&\makecell[c]{MMD$\downarrow$}
                &\makecell[c]{TBK$\uparrow$}\\
				
				\Xhline{1pt}
				
				R2DM \cite{nakashima2024lidar}
				&\makecell[c]{10000}
				&\makecell[c]{34149}      
				&\makecell[c]{86.79}
				&\makecell[c]{91.12}
				&\makecell[c]{0.42}
				&\makecell[c]{5.03}
                &\makecell[c]{21.80}\\
				
				Text2LiDAR \cite{wu2024text2lidar}
				&\makecell[c]{10000}
				&\makecell[c]{34149}      
				&\makecell[c]{85.12}
				&\makecell[c]{88.94}
				&\makecell[c]{0.37}
				&\makecell[c]{3.95}
                &\makecell[c]{23.41}\\

                T2LDM \cite{qu2026self}
				&\makecell[c]{10000}
				&\makecell[c]{34149}        
				&\makecell[c]{63.98}
				&\makecell[c]{65.79}
				&\makecell[c]{0.28}
				&\makecell[c]{3.03}
                &\makecell[c]{32.57}\\

				\cellcolor[rgb]{0.8549, 0.9098, 0.9882} T2LDM++ 
				&\cellcolor[rgb]{0.8549, 0.9098, 0.9882} \makecell[c]{10000}
				&\cellcolor[rgb]{0.8549, 0.9098, 0.9882} \makecell[c]{34149}     
				&\cellcolor[rgb]{0.8549, 0.9098, 0.9882} \makecell[c]{58.96}
				&\cellcolor[rgb]{0.8549, 0.9098, 0.9882} \makecell[c]{62.30}
				&\cellcolor[rgb]{0.8549, 0.9098, 0.9882} \makecell[c]{0.26}
				&\cellcolor[rgb]{0.8549, 0.9098, 0.9882} \makecell[c]{3.02}		&\cellcolor[rgb]{0.8549, 0.9098, 0.9882} \makecell[c]{34.17}\\

                \cellcolor[rgb]{0.8549, 0.9098, 0.9882} T2LDM++(ZS) 
				&\cellcolor[rgb]{0.8549, 0.9098, 0.9882} \makecell[c]{10000}
				&\cellcolor[rgb]{0.8549, 0.9098, 0.9882} \makecell[c]{34149}     
				&\cellcolor[rgb]{0.8549, 0.9098, 0.9882} \makecell[c]{60.73}
				&\cellcolor[rgb]{0.8549, 0.9098, 0.9882} \makecell[c]{62.14}
				&\cellcolor[rgb]{0.8549, 0.9098, 0.9882} \makecell[c]{0.27}
				&\cellcolor[rgb]{0.8549, 0.9098, 0.9882} \makecell[c]{3.03}		&\cellcolor[rgb]{0.8549, 0.9098, 0.9882} \makecell[c]{36.32}\\
                
				\Xhline{1pt}
				
			\end{tabular}
		}

    \caption{The text-guided results on T2nuScenes++. 'T2LDM++(ZS)' means the Zero-Shot Text-to-LiDAR generation results of T2LDM++ conditioned on 3D Boxes. T2LDM++ exhibits outstanding performance in  controllability and generation quality.}
    \label{tab631}
    \vspace{-15pt}
\end{table}

\begin{table}[h]
	\resizebox{0.475\textwidth}{!}{
			\begin{tabular}{p{2.5cm}|p{0.9cm}p{0.9cm}|p{1.0cm}p{1.1cm}p{1.0cm}p{1.0cm}p{1.0cm}}	
				\Xhline{1pt}
				
				{Methods}
				&\makecell[c]{GS}
				&\makecell[c]{RS}
				&\makecell[c]{FSVD$\downarrow$}
				&\makecell[c]{FPVD$\downarrow$}
				&\makecell[c]{JSD$\downarrow$}
				&\makecell[c]{MMD$\downarrow$}
                &\makecell[c]{TSK$\uparrow$}\\
				
				\Xhline{1pt}
				
				R2DM \cite{nakashima2024lidar}
				&\makecell[c]{10000}
				&\makecell[c]{23021}      
				&\makecell[c]{65.88}
				&\makecell[c]{63.15}
				&\makecell[c]{0.39}
				&\makecell[c]{5.05}
                &\makecell[c]{11.53}\\
				
				Text2LiDAR \cite{wu2024text2lidar}
				&\makecell[c]{10000}
				&\makecell[c]{23021}      
				&\makecell[c]{73.13}
				&\makecell[c]{68.49}
				&\makecell[c]{0.41}
				&\makecell[c]{5.23}
                &\makecell[c]{13.66}\\

                T2LDM \cite{qu2026self}
				&\makecell[c]{10000}
				&\makecell[c]{23021}        
				&\makecell[c]{29.42}
				&\makecell[c]{32.57}
				&\makecell[c]{0.34}
				&\makecell[c]{4.04}
                &\makecell[c]{19.95}\\

				\cellcolor[rgb]{1.0, 0.9020, 0.8} T2LDM++ 
				&\cellcolor[rgb]{1.0, 0.9020, 0.8} \makecell[c]{10000}
				&\cellcolor[rgb]{1.0, 0.9020, 0.8} \makecell[c]{23021}     
				&\cellcolor[rgb]{1.0, 0.9020, 0.8} \makecell[c]{27.59}
				&\cellcolor[rgb]{1.0, 0.9020, 0.8} \makecell[c]{30.58}
				&\cellcolor[rgb]{1.0, 0.9020, 0.8} \makecell[c]{0.31}
				&\cellcolor[rgb]{1.0, 0.9020, 0.8} \makecell[c]{3.87}		&\cellcolor[rgb]{1.0, 0.9020, 0.8} \makecell[c]{21.45}\\

                \cellcolor[rgb]{1.0, 0.9020, 0.8} T2LDM++(ZS) 
				&\cellcolor[rgb]{1.0, 0.9020, 0.8} \makecell[c]{10000}
				&\cellcolor[rgb]{1.0, 0.9020, 0.8} \makecell[c]{23021}     
				&\cellcolor[rgb]{1.0, 0.9020, 0.8} \makecell[c]{22.14}
				&\cellcolor[rgb]{1.0, 0.9020, 0.8} \makecell[c]{26.21}
				&\cellcolor[rgb]{1.0, 0.9020, 0.8} \makecell[c]{0.30}
				&\cellcolor[rgb]{1.0, 0.9020, 0.8} \makecell[c]{3.82}		&\cellcolor[rgb]{1.0, 0.9020, 0.8} \makecell[c]{25.78}\\
                
				\Xhline{1pt}
				
			\end{tabular}
		}

    \caption{The text-guided results on T2SemanticKITTI. 'T2LDM++(ZS)' means the Zero-Shot Text-to-LiDAR generation results of T2LDM++ conditioned on Semantic Labels. T2LDM++ achieves significantly improved text-guided generation results.}
    \label{tab632}
    \vspace{-20pt}
\end{table}

\subsection{Other Conditional Generation}

\label{sec64}

\begin{figure}[!t]
	\centering
	\includegraphics[width=0.48\textwidth]{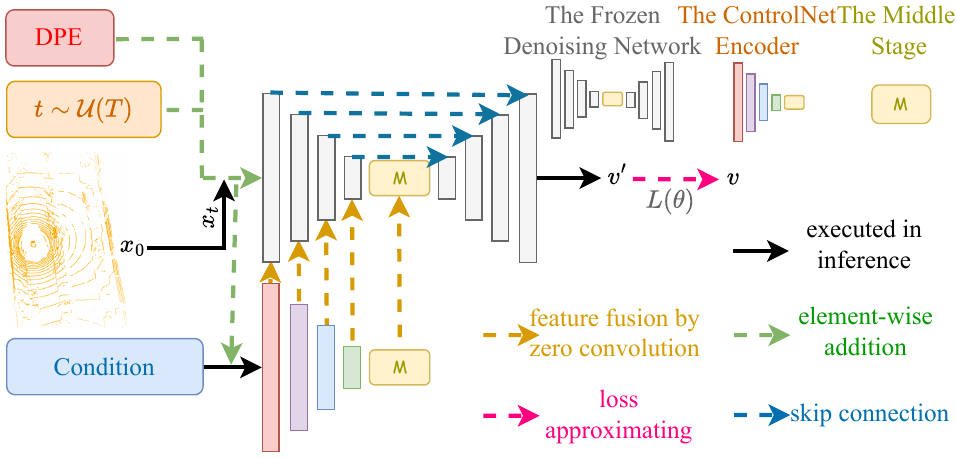}
	\caption{The overall framework of non-latent ControlNet \cite{qu2026self}. T2LDM++ is based on a U-Net DDPM rather than a DiT \cite{peebles2023scalable}, thus we employ an additional Encoder ($\mathcal{E}_c$) for conditional control after removing GN and freezing DN.}
	\label{fig16}
	
\end{figure}

\textbf{Non-latent ControlNet.} Following ControlNet \cite{zhang2023adding}, T2LDM++ can be extended to various conditional generation tasks by learning a Control Encoder $\mathcal{E}_c$ with a frozen unconditional DN, as illustrated in Fig.~\ref{fig16}. This is also the first exploration of ControlNet on non-latent DDPMs for 3D generation. 

\begin{figure*}[htp]
	\centering
	\includegraphics[width=0.99\textwidth]{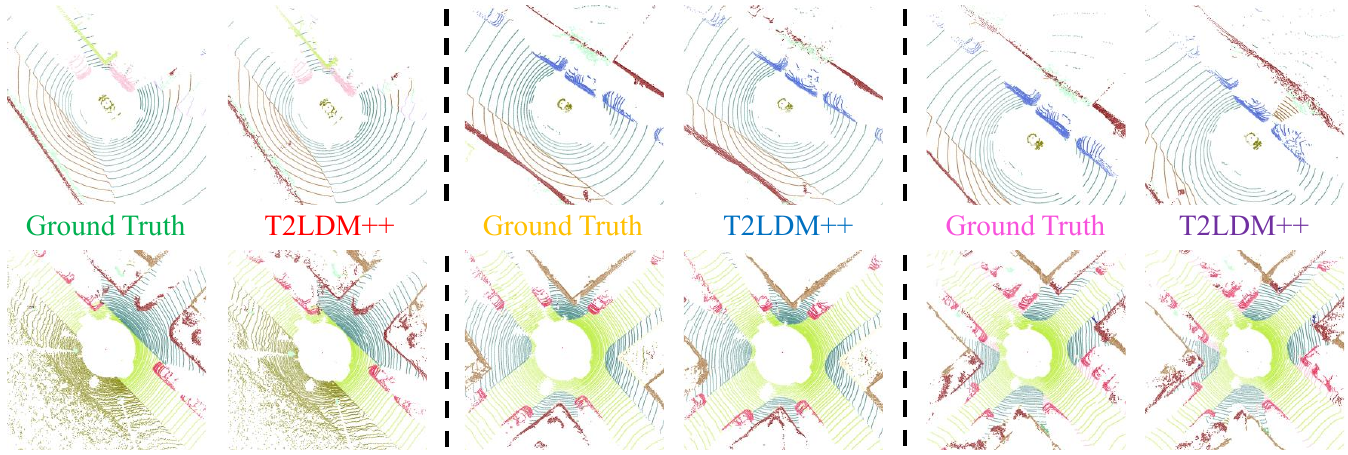}
	\caption{Semantic-to-LiDAR results on nuScenes (the top row) and SemanticKITTI (the bottom row).}
	\label{fig17}
	
\end{figure*}

\textbf{Semantic-to-LiDAR.} We first achieve Semantic-to-LiDAR (S2L) generation. We normalize the Semantic Map $ \in \mathbb{R}^{H \times W \times 1}$ to [0,1] via Semantic Label/Class Num, and then feed this into $\mathcal{E}_c$ as the control condition. Tab.~\ref{tab641} presents the generation results of S2L. Meanwhile, Fig.~\ref{fig17} visualizes the S2L generation results of T2LDM++. Furthermore, as discussed in Sec.~\ref{sec63}, S2L generation enables Zero-Shot Text-to-LiDAR generation on T2SemanticKITTI.

\begin{figure*}[htp]
	\centering
	\includegraphics[width=0.99\textwidth]{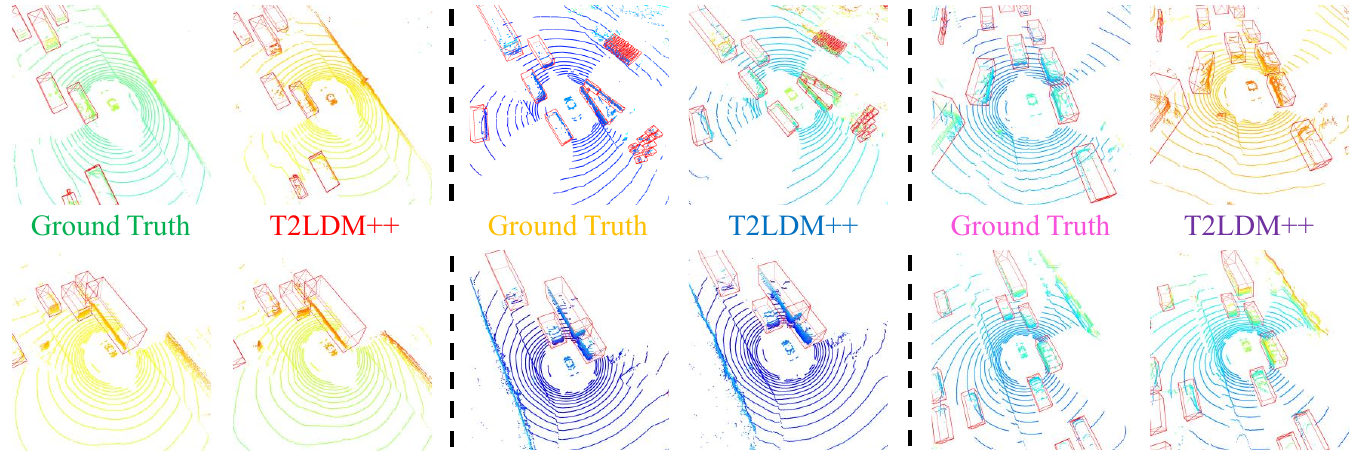}
	\caption{Box-to-LiDAR results on nuScenes. Semantic Maps bridge generation gap between 3D Boxes and LiDAR scenes.}
	\label{fig18}
	
\end{figure*}

\textbf{Box-to-LiDAR.} We also implement Box-to-LiDAR (B2L) generation. The 3D Box  $ \in \mathbb{R}^{M \times 7}$ cannot be directly encoded into feature maps aligned with the RM dimensions, thus we first convert this into the Semantic Map  $ \in \mathbb{R}^{H \times W \times 1}$ as the control condition. As shown in Tab.~\ref{tab642}, T2LDM++ achieves superior B2L generation. Meanwhile, Fig.~\ref{fig18} further demonstrates the effective generation results of T2LDM++. We observe that the B2L generation performs slightly below the S2L generation (see Tab.~\ref{tab631} and  Tab.~\ref{tab641}). This is because the Semantic Map is converted from the 3D Box contains a large amount of background labels and cannot provide scene structures as finely as the original Semantic Label. This means that \textit{the effectiveness of ControlNet is positively correlated with the input condition for the coverage ratio of the generation target}.

\begin{figure*}[htp]
	\centering
	\includegraphics[width=0.99\textwidth]{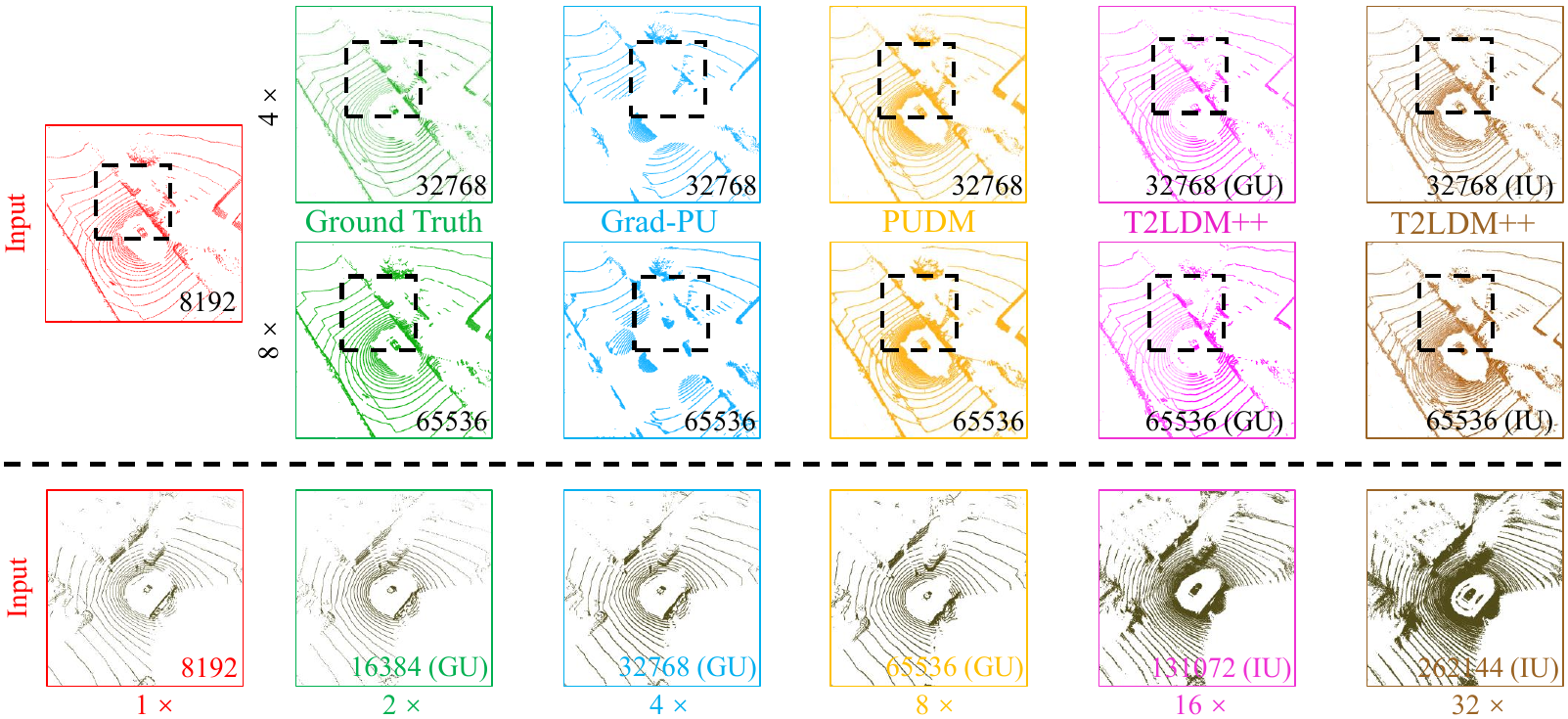}
	\caption{Sparse-to-Dense results on nuScenes. 'GU' means the complete Generation Upsampling. 'IU' denotes the use of interpolation guidance in inference \cite{qu2024conditional}. T2LDM++ can support arbitrary upsampling rates (the bottom row).}
	\label{fig19}
	
\end{figure*}

\begin{figure*}[htp]
	\centering
	\includegraphics[width=0.99\textwidth]{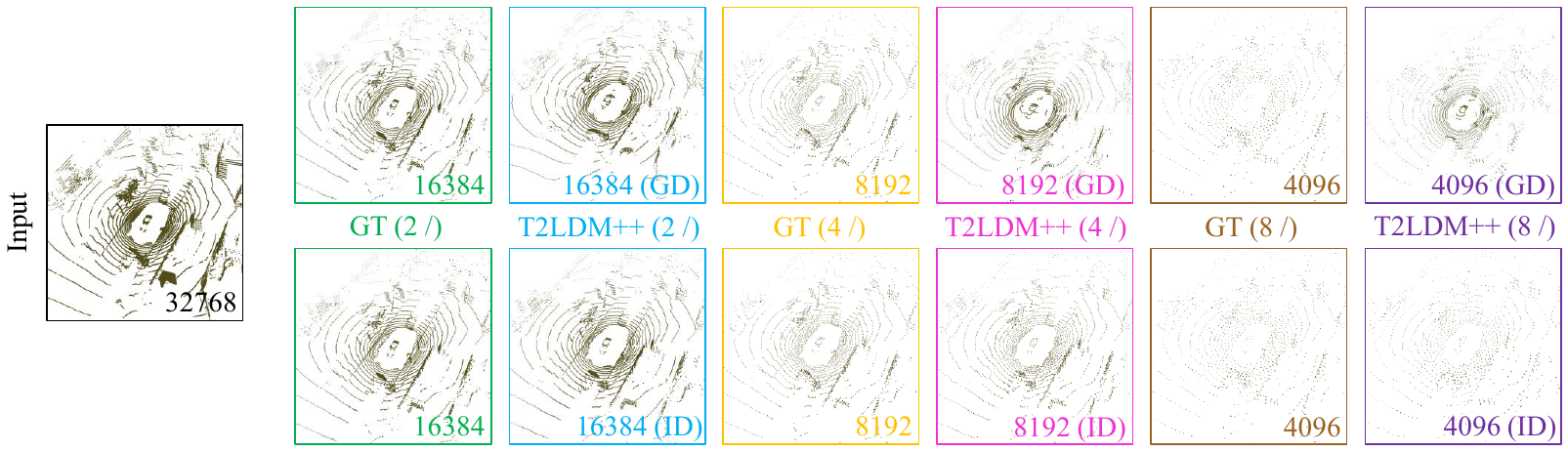}
	\caption{Dense-to-Sparse results on nuScenes. 'GD' means the complete Generation Downsampling. 'ID' denotes the use of interpolation guidance in inference \cite{qu2024conditional}. T2LDM++ can achieve downsampling at arbitrary rates without retraining.}
	\label{fig20}
	
\end{figure*}

\begin{figure*}[t]
    \centering
    \includegraphics[width=0.99\textwidth]{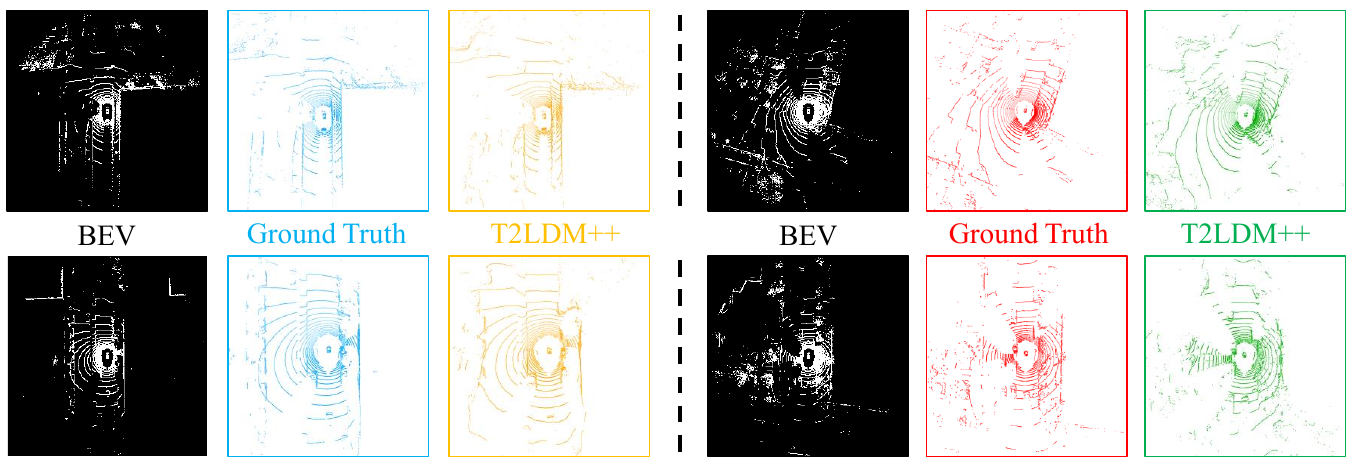}

    \vspace{8mm}

    \includegraphics[width=0.99\textwidth]{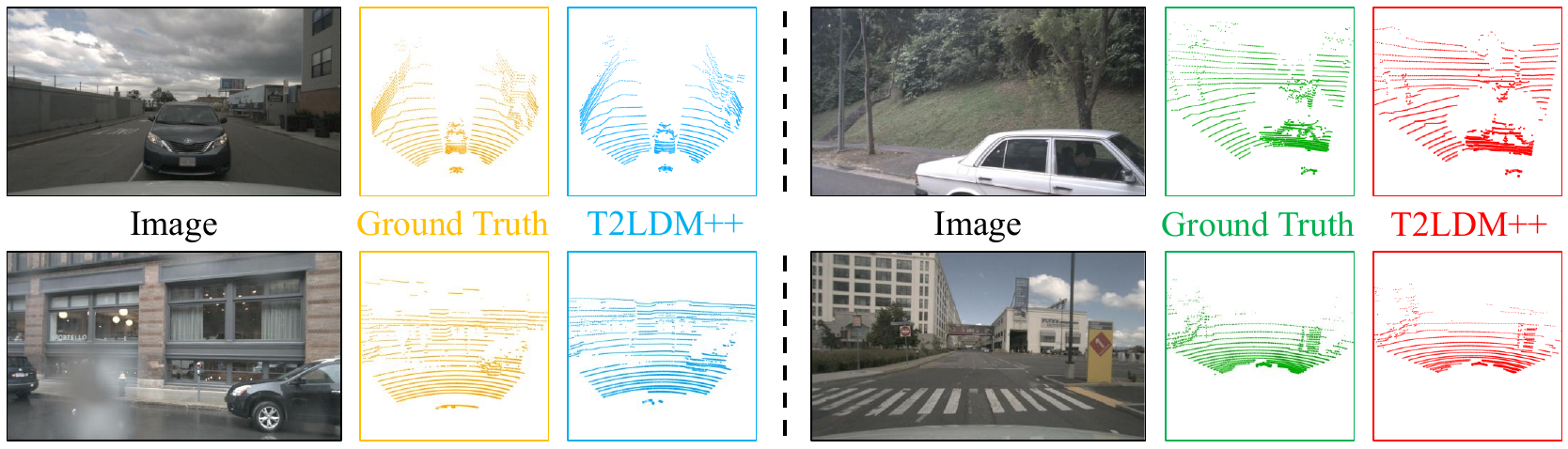}

    \caption{
    BEV-to-LiDAR and Camera-to-LiDAR results on nuScenes.
    Top: BEV-to-LiDAR results.
    Bottom: Camera-to-LiDAR results. T2LDM++ can effectively generate full and partial LiDAR scenes from binary maps and camera images.
    }
    \label{fig21_22}
\end{figure*}

	

	

\begin{table}[!t]
	\resizebox{0.475\textwidth}{!}{
			\begin{tabular}{p{2.5cm}|p{0.9cm}p{0.9cm}|p{1.0cm}p{1.1cm}p{1.0cm}p{1.0cm}}	
				\Xhline{1pt}
				
				{Methods}
				&\makecell[c]{GS}
				&\makecell[c]{RS}
				&\makecell[c]{FSVD$\downarrow$}
				&\makecell[c]{FPVD$\downarrow$}
				&\makecell[c]{JSD$\downarrow$}
				&\makecell[c]{MMD$\downarrow$}\\
				
				\Xhline{1pt}
				
				(Un)T2LDM++
				&\makecell[c]{10000}
				&\makecell[c]{34149}      
				&\makecell[c]{58.68}
				&\makecell[c]{61.09}
				&\makecell[c]{0.25}
				&\makecell[c]{3.00}\\

				\cellcolor[rgb]{0.9725, 0.8078, 0.8}
                (S2L)T2LDM++
				&\cellcolor[rgb]{0.9725, 0.8078, 0.8}
                \makecell[c]{10000}
				&\cellcolor[rgb]{0.9725, 0.8078, 0.8}
                \makecell[c]{34149}        
				&\cellcolor[rgb]{0.9725, 0.8078, 0.8}
                \makecell[c]{58.72}
				&\cellcolor[rgb]{0.9725, 0.8078, 0.8}
                \makecell[c]{60.89}
				&\cellcolor[rgb]{0.9725, 0.8078, 0.8}
                \makecell[c]{0.25}
				&\cellcolor[rgb]{0.9725, 0.8078, 0.8}
                \makecell[c]{3.00}\\
                
                \Xhline{1pt}
            
                (Un)T2LDM++
				&\makecell[c]{10000}
				&\makecell[c]{23021}        
				&\makecell[c]{24.37}
				&\makecell[c]{28.14}
				&\makecell[c]{0.31}
				&\makecell[c]{3.85}\\

				\cellcolor[rgb]{0.9725, 0.8078, 0.8} (S2L)T2LDM++ 
				&\cellcolor[rgb]{0.9725, 0.8078, 0.8}
                \makecell[c]{10000}
				&\cellcolor[rgb]{0.9725, 0.8078, 0.8}
                \makecell[c]{23021}        
				&\cellcolor[rgb]{0.9725, 0.8078, 0.8}
                \makecell[c]{22.25}
				&\cellcolor[rgb]{0.9725, 0.8078, 0.8}
                \makecell[c]{26.38}
				&\cellcolor[rgb]{0.9725, 0.8078, 0.8}
                \makecell[c]{0.30}
				&\cellcolor[rgb]{0.9725, 0.8078, 0.8}
                \makecell[c]{3.82}\\
                
				\Xhline{1pt}
				
			\end{tabular}
		}

    \caption{The Semantic-to-LiDAR results on nuScenes (the top rows) and SemanticKITTI (the bottom rows).}
    \label{tab641}
    \vspace{-10pt}
\end{table}


\begin{table}[!t]
	\resizebox{0.475\textwidth}{!}{
			\begin{tabular}{p{2.5cm}|p{0.9cm}p{0.9cm}|p{1.0cm}p{1.1cm}p{1.0cm}p{1.0cm}}	
				\Xhline{1pt}
				
				{Methods}
				&\makecell[c]{GS}
				&\makecell[c]{RS}
				&\makecell[c]{FSVD$\downarrow$}
				&\makecell[c]{FPVD$\downarrow$}
				&\makecell[c]{JSD$\downarrow$}
				&\makecell[c]{MMD$\downarrow$}\\
				
				\Xhline{1pt}
				
				(Un)T2LDM++
				&\makecell[c]{10000}
				&\makecell[c]{34149}      
				&\makecell[c]{58.68}
				&\makecell[c]{61.09}
				&\makecell[c]{0.25}
				&\makecell[c]{3.00}\\

				\cellcolor[rgb]{1.0, 0.9490, 0.8}
                (B2L)T2LDM++
				&\cellcolor[rgb]{1.0, 0.9490, 0.8}
                \makecell[c]{10000}
				&\cellcolor[rgb]{1.0, 0.9490, 0.8}
                \makecell[c]{34149}        
				&\cellcolor[rgb]{1.0, 0.9490, 0.8}
                \makecell[c]{60.80}
				&\cellcolor[rgb]{1.0, 0.9490, 0.8}
                \makecell[c]{62.20}
				&\cellcolor[rgb]{1.0, 0.9490, 0.8}
                \makecell[c]{0.26}
				&\cellcolor[rgb]{1.0, 0.9490, 0.8}
                \makecell[c]{3.01}\\
                
				\Xhline{1pt}
				
			\end{tabular}
		}

    \caption{The Box-to-LiDAR results on nuScenes. By introducing Semantic Labels as an intermediate representation, B2L generation can be achieved.}
    \label{tab642}
    \vspace{-10pt}
\end{table}


\begin{table}[!t]
	\scriptsize
	\resizebox{0.48\textwidth}{!}{
		\begin{tabular}{p{2.5cm}p{1.0cm}p{1.0cm}p{1.0cm}p{0.005cm}p{1.0cm}p{1.0cm}p{1.0cm}}	
			\Xhline{1pt}
			
			\makecell[l]{\multirow{2}{*}{Methods}}
			&\multicolumn{3}{c}{$4\times$} 
			&\quad
			&\multicolumn{3}{c}{$8\times$} \\
			
			&\makecell[c]{CD$\downarrow$}
			&\makecell[c]{MSE$\downarrow$}
			&\makecell[c]{EMD$\downarrow$}
			&\quad
			&\makecell[c]{CD$\downarrow$}
			&\makecell[c]{MSE$\downarrow$}
			&\makecell[c]{EMD$\downarrow$}\\
			\hline
			
			\makecell[l]{Grad-PU \cite{he2023grad}}
			&\makecell[c]{0.400}
			&\makecell[c]{4.169}
			&\makecell[c]{2.324}
			&\quad
			&\makecell[c]{0.364}
			&\makecell[c]{4.031}
			&\makecell[c]{2.142}\\
			
			\makecell[l]{PUDM \cite{qu2024conditional}}
			&\makecell[c]{0.198}
			&\makecell[c]{4.275}
			&\makecell[c]{2.124}
			&\quad
			&\makecell[c]{0.103}
			&\makecell[c]{4.102}
			&\makecell[c]{1.914}\\

			\cellcolor[rgb]{0.8352, 0.9098, 0.8314} \makecell[l]{(S2D)T2LDM++}
			&\cellcolor[rgb]{0.8352, 0.9098, 0.8314} \makecell[c]{0.097}
			&\cellcolor[rgb]{0.8352, 0.9098, 0.8314} \makecell[c]{3.594}
			&\cellcolor[rgb]{0.8352, 0.9098, 0.8314} \makecell[c]{1.974}
			&\cellcolor[rgb]{0.8352, 0.9098, 0.8314} \quad
			&\cellcolor[rgb]{0.8352, 0.9098, 0.8314} \makecell[c]{0.069}
			&\cellcolor[rgb]{0.8352, 0.9098, 0.8314} \makecell[c]{3.559}
			&\cellcolor[rgb]{0.8352, 0.9098, 0.8314} \makecell[c]{1.898}\\
            
			\Xhline{1pt}
			
		\end{tabular}
	}
	\caption{The results of the $4 \times$ rate and the $8 \times$ rate on nuScenes. Using non-latent ControlNet, T2LDM++ can achieve significantly upsampling results.}
	\label{tab643}
    \vspace{-10pt}
\end{table}

\begin{table}[!t]
	\resizebox{0.475\textwidth}{!}{
			\begin{tabular}{p{2.5cm}|p{0.9cm}p{0.9cm}|p{1.0cm}p{1.1cm}p{1.0cm}p{1.0cm}}	
				\Xhline{1pt}
				
				{Methods}
				&\makecell[c]{GS}
				&\makecell[c]{RS}
				&\makecell[c]{FSVD$\downarrow$}
				&\makecell[c]{FPVD$\downarrow$}
				&\makecell[c]{JSD$\downarrow$}
				&\makecell[c]{MMD$\downarrow$}\\
				
				\Xhline{1pt}
				
				(Un)T2LDM++
				&\makecell[c]{10000}
				&\makecell[c]{34149}      
				&\makecell[c]{58.68}
				&\makecell[c]{61.09}
				&\makecell[c]{0.25}
				&\makecell[c]{3.00}\\

				\cellcolor[rgb]{0.9922, 0.8275, 0.9647} (V2L)T2LDM++
				&\cellcolor[rgb]{0.9922, 0.8275, 0.9647} \makecell[c]{10000}
				&\cellcolor[rgb]{0.9922, 0.8275, 0.9647} \makecell[c]{34149}        
				&\cellcolor[rgb]{0.9922, 0.8275, 0.9647} \makecell[c]{58.17}
				&\cellcolor[rgb]{0.9922, 0.8275, 0.9647} \makecell[c]{60.32}
				&\cellcolor[rgb]{0.9922, 0.8275, 0.9647} \makecell[c]{0.25}
				&\cellcolor[rgb]{0.9922, 0.8275, 0.9647} \makecell[c]{3.00}\\
                
				\Xhline{1pt}
				
			\end{tabular}
		}

    \caption{The BEV-to-LiDAR results on nuScenes. T2LDM++ enables LiDAR scene generation conditioned on binary maps.}
    \label{tab644}
    \vspace{-15pt}
\end{table}

\textbf{Sparse-to-Dense.} Meanwhile, we perform Sparse-to-Dense (S2D) generation. Following the official nuScenes split, we downsample the LiDAR point clouds by $4 \times$ using FPS, to generate sparse and dense point clouds for training (28,140 samples). For testing, 6,019 samples are downsampled by $4 \times$ as sparse point clouds, while the original and $2 \times$ upsampled point clouds are used as the $4 \times$ and $8 \times$ Ground Truths, respectively. For condition adaptation, we encode the sparse LiDAR into the same feature space as $\mathrm{RM} \in \mathbb{R}^{H \times W \times 1}$ as the input of $\mathcal{E}_c$. We follow existing methods  \cite{he2023grad, qu2024conditional} by directly validating the PU-GAN \cite{li2019pu} pretrained model on nuScenes. For fair comparison, we unify the scale by normalizing the point coordinates to [0,1] \cite{qu2026self}. As shown in Tab.~\ref{tab643}, T2LDM++ achieves superior upsampling results. Meanwhile, Fig.~\ref{fig19} presents the qualitative results.

\textbf{Dense-to-Sparse.} Furthermore, since the output LiDAR point cloud shape is determined by the input noise \cite{qu2024conditional, qu2026self}, T2LDM++, capable of upsampling, can directly achieve downsampling. Fig.~\ref{fig20} presents the Dense-to-Sparse (D2S) generation results at arbitrary rates without retraining. 

\textbf{BEV-to-LiDAR.} Moreover, we also implement BEV-to-LiDAR (V2L) generation for T2LDM++. This first converts LiDAR data into the Bird’s Eye View. $\mathrm{BEV} \in \mathbb{R}^{256 \times 256 \times 1}$, containing binary pixel values (0 (empty) or 1 (occupied)), replicated three times along the single channel dimension, is used as the input to $\mathcal{E}_c$. Meanwhile, to adapt to $\mathrm{RM} \in \mathbb{R}^{H \times W \times 2}$, we use the first two layers \cite{huang2022imfnet} of a pretrained ResNet-34 \cite{he2016deep} to extract image features $\mathrm{I} \in \mathbb{R}^{32 \times 32 \times 3}$ from $\mathrm{BEV} \in \mathbb{R}^{256 \times 256 \times 3}$. Subsequently, $\mathrm{I}$ is resized to (H,W) via bilinear interpolation. Tab.~\ref{tab644} presents V2L generation results of T2LDM++. Furthermore, the visualization is shown in Fig.~\ref{fig21_22}(top).

\textbf{Camera-to-LiDAR.} Finally, we further validate T2LDM++ on Camera-to-LiDAR (C2L) generation. To better align the camera data with the LiDAR scene, we compute the angular range of each image for LiDAR scene according to the camera parameters. This divides the LiDAR scene into six parts, corresponding to “CAM$\_$FRONT”, “CAM$\_$FRONT$\_$LEFT”, “CAM$\_$FRONT$\_$RIGHT”, “CAM$\_$BACK”, “CAM$\_$BACK$\_$LEFT”, and “CAM$\_$BACK$\_$RIGHT” from nuScenes, respectively. Subsequently, based on this setting, we retrain T2LDM++ for unconditional generation using angle-segmented LiDAR partial scenes. Furthermore, the Camera Image (CI) is fed into $\mathcal{E}_c$ as the control condition to achieve C2L generation. Similarly, as the dimensions of CI $ \in \mathbb{R}^{h \times w \times 3}$ are incompatible with   $\mathrm{RM} \in \mathbb{R}^{H \times W \times 2}$, CI is processed in the same way as the BEV image. Fig.~\ref{fig21_22}(bottom) further illustrates the visualization results. In fact, we find that directly using the DN trained on full scene generation leads to very poor C2L results. This further validates the previous conclusion regarding the effectiveness of ControlNet in Box-to-LiDAR generation.




\subsection{Ablation Study}

To evaluate the effectiveness of each component, we first conduct ablation studies on T2LDM++. Then, we analyze the suitability of different sampling strategies for T2LDM++. Finally, we evaluate TBK across multiple detection models to verify the sensitivity of the result.

\begin{table}[h]
	\resizebox{0.48\textwidth}{!}{
			\begin{tabular}{p{2.5cm}|p{0.9cm}p{0.9cm}|p{1.0cm}p{1.1cm}p{1.0cm}p{1.0cm}}	
				\Xhline{1pt}
				
				{Methods}
				&\makecell[c]{GS}
				&\makecell[c]{RS}
				&\makecell[c]{FSVD$\downarrow$}
				&\makecell[c]{FPVD$\downarrow$}
				&\makecell[c]{JSD$\downarrow$}
				&\makecell[c]{MMD$\downarrow$}\\
				
				\Xhline{1pt}
				
				T2LDM++$^{\emptyset}$
				&\makecell[c]{10000}
				&\makecell[c]{34149}      
				&\makecell[c]{67.88}
				&\makecell[c]{69.07}
				&\makecell[c]{0.27}
				&\makecell[c]{3.03}\\
				
				T2LDM++$^{D}$ 
				&\makecell[c]{10000}
				&\makecell[c]{34149}      
				&\makecell[c]{65.25}
				&\makecell[c]{67.46}
				&\makecell[c]{0.26}
				&\makecell[c]{3.01}\\
				
				T2LDM++$^{S}$ 
				&\makecell[c]{10000}
				&\makecell[c]{34149}      
				&\makecell[c]{61.55}
				&\makecell[c]{63.18}
				&\makecell[c]{0.25}
				&\makecell[c]{3.01}\\
				
				\cellcolor[rgb]{0.9725, 0.8078, 0.8} T2LDM++ 
				&\cellcolor[rgb]{0.9725, 0.8078, 0.8}
                \makecell[c]{10000}
				&\cellcolor[rgb]{0.9725, 0.8078, 0.8}
                \makecell[c]{34149}   
				&\cellcolor[rgb]{0.9725, 0.8078, 0.8}
                \makecell[c]{58.68}
				&\cellcolor[rgb]{0.9725, 0.8078, 0.8}
                \makecell[c]{61.09}
				&\cellcolor[rgb]{0.9725, 0.8078, 0.8}
                \makecell[c]{0.25}
				&\cellcolor[rgb]{0.9725, 0.8078, 0.8}
                \makecell[c]{3.00}\\
				
				\Xhline{1pt}
				
			\end{tabular}
		}
		\caption{Ablation study of component effectiveness on unconditional nuScenes generation. T2LDM++$^{\emptyset}$, T2LDM++$^{D}$, and T2LDM++$^{S}$ denote removing DPE and SCRG, only retaining DPE, and only retaining SCRG, respectively. The directional priors and  geometry-aware regularization provided by DPE and SCRG  significantly improve the generation performance of T2LDM++.}
		\label{tab651}
        \vspace{-15pt}
\end{table}

\begin{table}[h]
	\resizebox{0.475\textwidth}{!}{
			\begin{tabular}{p{2.5cm}|p{1.0cm}p{1.0cm}p{1.0cm}|p{1.1cm}p{1.1cm}p{1.2cm}p{1.1cm}}	
				\Xhline{1pt}
				
				{Methods (30k)}
				&\makecell[c]{IP}
				&\makecell[c]{Steps}
				&\makecell[c]{GS}
				&\makecell[c]{FSVD$\downarrow$}
				&\makecell[c]{FPVD$\downarrow$}
				&\makecell[c]{JSD$\downarrow$}
				&\makecell[c]{MMD$\downarrow$}\\
				
				\Xhline{1pt}
				
				R2DM \cite{nakashima2024lidar}
				&\makecell[c]{31.1M}
				&\makecell[c]{1024}
				&\makecell[c]{10000}      
				&\makecell[c]{173.14}
				&\makecell[c]{148.55}
				&\makecell[c]{0.51}
				&\makecell[c]{9.97}\\
				
				Text2LiDAR \cite{wu2024text2lidar}
				&\makecell[c]{45.8M}  
				&\makecell[c]{1024}
				&\makecell[c]{10000}      
				&\makecell[c]{342.24}
				&\makecell[c]{323.15}
				&\makecell[c]{0.86}
				&\makecell[c]{17.13}\\
				
				T2LDM++$^{D}$
				&\makecell[c]{30.6M}  
				&\makecell[c]{1024}
				&\makecell[c]{10000}      
				&\makecell[c]{87.94}
				&\makecell[c]{88.44}
				&\makecell[c]{0.43}
				&\makecell[c]{8.41}\\
				
				\cellcolor[rgb]{1.0, 0.9490, 0.8}
                T2LDM++ 
				&\cellcolor[rgb]{1.0, 0.9490, 0.8}
                \makecell[c]{30.6M}
				&\cellcolor[rgb]{1.0, 0.9490, 0.8}
                \makecell[c]{1024}
				&\cellcolor[rgb]{1.0, 0.9490, 0.8}
                \makecell[c]{10000}        
				&\cellcolor[rgb]{1.0, 0.9490, 0.8}
                \makecell[c]{41.93}
				&\cellcolor[rgb]{1.0, 0.9490, 0.8}
                \makecell[c]{48.75}
				&\cellcolor[rgb]{1.0, 0.9490, 0.8}
                \makecell[c]{0.33}
				&\cellcolor[rgb]{1.0, 0.9490, 0.8}
                \makecell[c]{4.50}\\
				
				\Xhline{1pt}
				
			\end{tabular}
		} 
		\caption{The results on KITTI-360 at 30k iterations. 'IP' means the Inference Parameters.  The learning rate is $1e^{-4}$ for all methods \cite{qu2026self}. SCRG enables DN to learn high-frequency details under geometry-aware regularization in the early stage of training.}
		\label{tab652}
        \vspace{-15pt}
\end{table}


\begin{table}[h]
	\resizebox{0.475\textwidth}{!}{
			\begin{tabular}{p{2.5cm}|p{0.9cm}p{0.9cm}|p{1.0cm}p{1.1cm}p{1.0cm}p{1.0cm}}	
				\Xhline{1pt}
				
				{Methods}
				&\makecell[c]{GS}
				&\makecell[c]{RS}
				&\makecell[c]{FSVD$\downarrow$}
				&\makecell[c]{FPVD$\downarrow$}
				&\makecell[c]{JSD$\downarrow$}
				&\makecell[c]{MMD$\downarrow$}\\
				
				\Xhline{1pt}

				Pretrained Mode
				&\makecell[c]{10000}
				&\makecell[c]{23021}      
				&\makecell[c]{31.01}
				&\makecell[c]{35.77}
				&\makecell[c]{0.35}
				&\makecell[c]{4.11}\\
                
				T2LDM++$^{D}$ 
				&\makecell[c]{10000}
				&\makecell[c]{23021}      
				&\makecell[c]{29.54}
				&\makecell[c]{33.16}
				&\makecell[c]{0.33}
				&\makecell[c]{3.98}\\

				Frozen SCRG 
				&\makecell[c]{10000}
				&\makecell[c]{23021}      
				&\makecell[c]{26.12}
				&\makecell[c]{30.31}
				&\makecell[c]{0.32}
				&\makecell[c]{3.90}\\
				
				\cellcolor[rgb]{0.8352, 0.9098, 0.8314} End-to-End Mode 
				&\cellcolor[rgb]{0.8352, 0.9098, 0.8314} \makecell[c]{10000}
				&\cellcolor[rgb]{0.8352, 0.9098, 0.8314} \makecell[c]{23021}        
				&\cellcolor[rgb]{0.8352, 0.9098, 0.8314} \makecell[c]{24.37}
				&\cellcolor[rgb]{0.8352, 0.9098, 0.8314} \makecell[c]{28.14}
				&\cellcolor[rgb]{0.8352, 0.9098, 0.8314} \makecell[c]{0.31}
				&\cellcolor[rgb]{0.8352, 0.9098, 0.8314} \makecell[c]{3.85}\\
				
				\Xhline{1pt}
				
			\end{tabular}
		}
		\caption{Ablation study of End-to-End vs. Pretrained training for SCRG on SemanticKITTI.}
		\label{tab653}
        \vspace{-15pt}
\end{table}


\textbf{Component Effectiveness.} We first validate the effectiveness of different components of T2LDM++ on nuScenes. As shown in Tab.~\ref{tab651}, the geometric regularization provided by SCRG and the real directional priors introduced by DPE can significantly improve the generation performance of T2LDM++, demonstrating the contribution of each component.

\begin{figure*}[htp]
	\centering
	\includegraphics[width=0.99\textwidth]{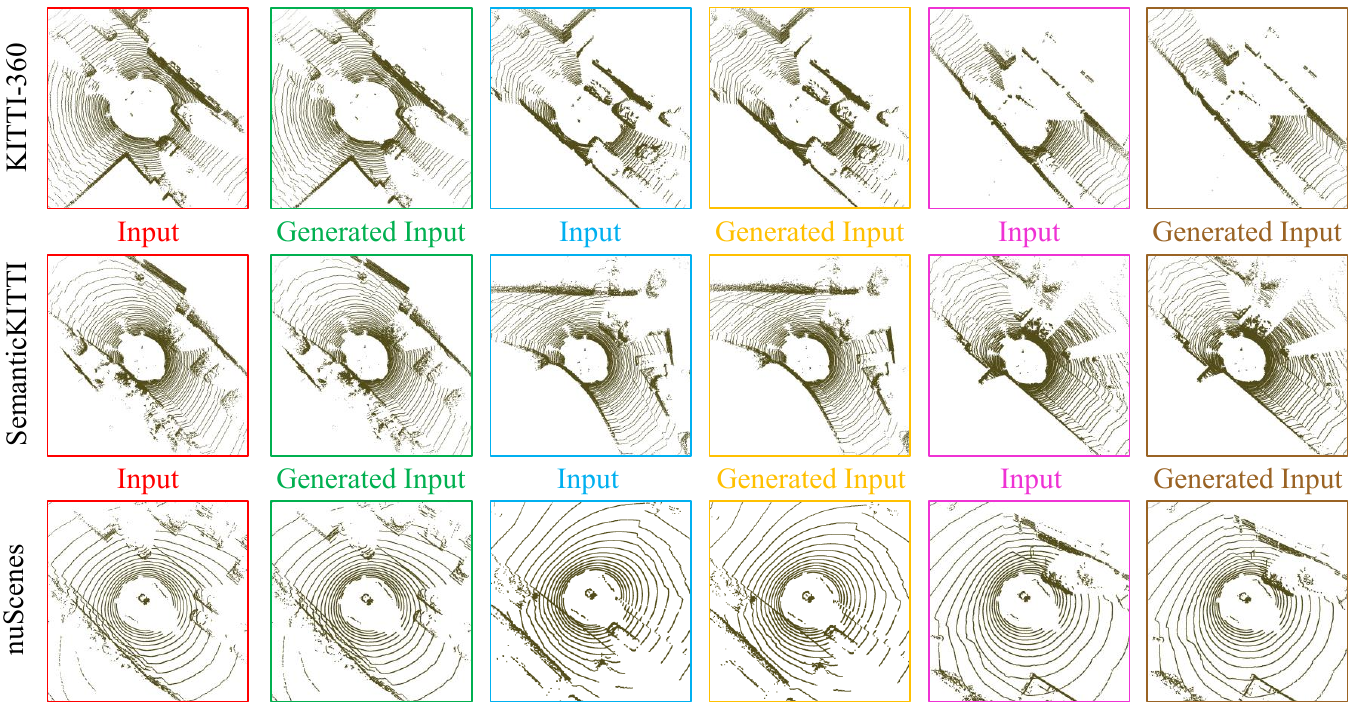}
	\caption{Reconstruction qualitative  results of GN. GN can effectively reconstruct the input, perceiving object details in scenes, providing geometry-aware regularization. This is consistent with the results in Fig.~\ref{fig8}.}
	\label{fig23}
	
\end{figure*}

\textbf{Convergence Speed.} Meanwhile, similar to the previous version \cite{qu2026self}, we further evaluate the effect of SCRG on the convergence speed of T2LDM++ under 30K iterations on KITTI-360. All methods are trained with a learning rate of $1e^{-4}$. Tab.~\ref{tab652} shows that SCRG enables DN to learn high-frequency reconstruction details via geometry-aware regularization in the early stages of training, thereby accelerating convergence. This remains consistent with the conclusion in Sec.~\ref{sec41} (see Fig.~\ref{fig7}(d)). Meanwhile, Fig.~\ref{fig23} presents the qualitative reconstruction results of GN.

\begin{figure*}[htp]
	\centering
	\includegraphics[width=0.99\textwidth]{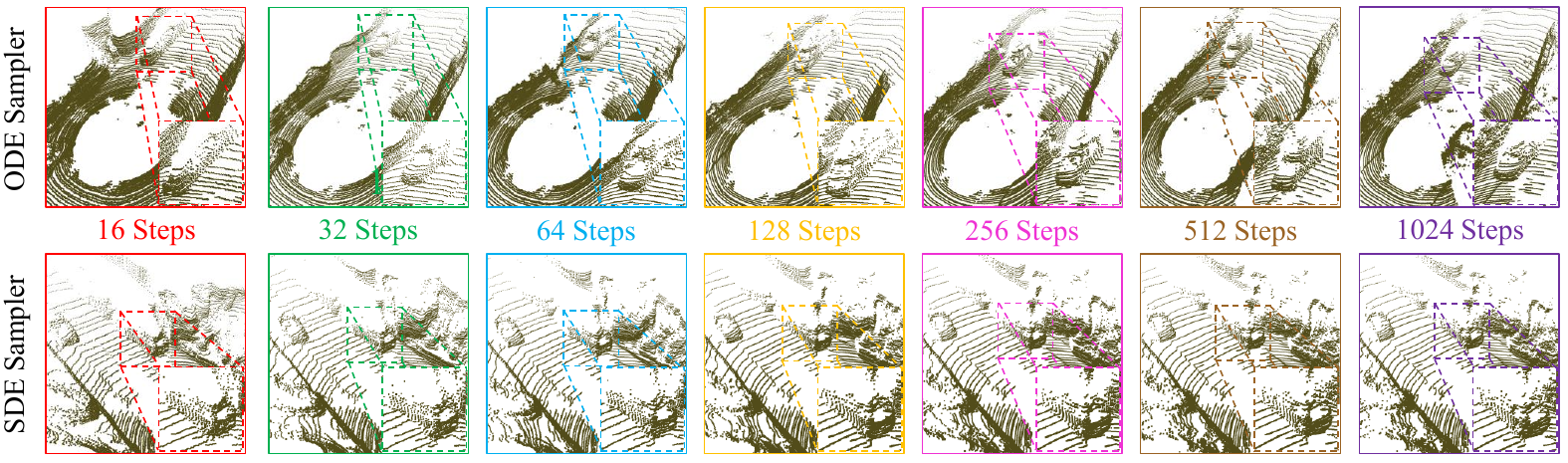}
	\caption{Visualization results of different samplers under different sampling steps for T2LDM++ on KITTI-360. The SDE sampler generates finer object details in the scene than the ODE sampler. The SDE sampler (64 steps) is recommended. }
	\label{fig24}
	
\end{figure*}

\textbf{End-to-End Mode vs. Pretrained Mode.} Furthermore, we also conduct the ablation study on different training strategies of SCRG on SemanticKITTI. Tab.~\ref{tab653} shows that the pretraining paradigm even leads to a performance drop. This is because GN cannot dynamically adapt to the evolving representations of DN, struggling to provide effective geometric regularization. This further validates the conclusion in Sec.~\ref{sec41} that T2LDM++ achieves optimal performance only when GN participates throughout the full training process (see Fig.~\ref{fig7}).

\textbf{Sampling Steps.} Moreover, we further analyze the sampling strategies for T2LDM++. SDE-based (DDPMs \cite{ho2020denoising}) and ODE-based (DDIMs \cite{song2020denoising}) sampling strategies are compared. In Tab.~\ref{tab654}, ODEs (64 steps) demonstrate a better trade-off between sampling steps and performance. However, we observe that SDEs produce more significant qualitative results, as shown in Fig.~\ref{fig24}. In fact, compared with regular images, raw point clouds acquired from 3D sensors are inherently sparse and noisy \cite{qi2017pointnet++, milioto2019rangenet++}. ODEs without stochasticity struggle to model the scattered distribution of points, leading to a lack of realism in generated scenes. In contrast, SDEs can introduce stochasticity to simulate distribution degeneration and local perturbations in sparse regions. Therefore, \textit{ODE-based sampling methods (DDIMs \cite{song2020denoising} or DPMs \cite{lu2022dpm}) may be more suitable for regular data, while SDE-based ones (DDPMs \cite{ho2020denoising}) may be more effective for irregular data}.

\textbf{Generalization for TBK.} Finally, we evaluate the sensitivity of TBK on different models. As shown in Tab.~\ref{tab655}, even models with significantly different detection performance (VoxelNeXt \cite{chen2023voxelnext} and FSHNet \cite{liu2025fshnet}) have a negligible influence on TBK results. This is because the TBK results mainly depend on \textit{the existence of target objects (whether "car" is detected), rather than the exact locations}. Therefore, this relaxes the performance requirement for the detector.

\begin{table}[h]
        \scriptsize
        \tabcolsep 2pt
        \resizebox{0.475\textwidth}{!}{
	\begin{tabular}{p{1.9cm}|p{2.0cm}|p{2.3cm}p{2.0cm}p{2.0cm}p{2.0cm}}	

        \Xhline{1pt}
        \makecell[l]{Methods}
        &\makecell[c]{GS}
        &\makecell[c]{1024($\sim$\textbf{\textcolor[rgb]{0.6039,0.3843,0.1451}{1348}}s)}
        &\makecell[c]{256($\sim$\textbf{\textcolor[rgb]{0.6039,0.3843,0.1451}{376}}s)}
        &\makecell[c]{64($\sim$\textbf{\textcolor[rgb]{0.6039,0.3843,0.1451}{89}}s)}
        &\makecell[c]{16($\sim$\textbf{\textcolor[rgb]{0.6039,0.3843,0.1451}{28}}s)}\\
        \Xhline{1pt}

        \makecell[l]{ODE Sam.}
        &\makecell[c]{10000}
        &\makecell[c]{29.37/32.15}
        &\makecell[c]{33.92/36.41}
        &\makecell[c]{31.01/34.78}
        &\makecell[c]{48.62/56.23}\\

        \makecell[l]{SDE Sam.}
        &\makecell[c]{10000}
        &\makecell[c]{23.34/26.86}
        &\makecell[c]{37.81/36.10}
        &\makecell[c]{38.44/37.25}
        &\makecell[c]{47.32/52.33}\\

        \Xhline{1pt}
	\end{tabular}
 }
	\caption{The results of multiple samplers on unconditional KITTI-360. $\sim$\textbf{\textcolor[rgb]{0.6039,0.3843,0.1451}{X}}s means the time of sampling 256 samples on 8 RTX 4090 GPUs (32 (batch size) $\times$ 8 (GPUs) = 256 samples). Meanwhile, 'XX/XX' indicates 'FSVD/FPVD'. ODEs (64 steps) exhibit the better trade-off between sampling steps and performance.}
	\label{tab654}
    \vspace{-30pt}
\end{table}


\begin{table}[h]
        \scriptsize
        \resizebox{0.475\textwidth}{!}{
	\begin{tabular}{p{2.7cm}|p{1.3cm}|p{1.5cm}p{1.3cm}p{1.3cm}p{2.0cm}p{1.3cm}}	

        \Xhline{1pt}
        \makecell[l]{Methods}
        &\makecell[c]{GS}
        &\makecell[c]{$\#$Params}
        &\makecell[c]{NDS$\uparrow$}
        &\makecell[c]{mAP$\uparrow$}
        &\makecell[c]{IMT/IMM$\downarrow$}
        &\makecell[c]{TBK$\uparrow$}\\
        \Xhline{1pt}

        \makecell[l]{VoxelNeXt\dag \cite{chen2023voxelnext}}
        &\makecell[c]{10000}
        &\makecell[c]{8.0M}
        &\makecell[c]{68.7}
        &\makecell[c]{63.5}
        &\makecell[c]{0.02s/0.6G}
        &\makecell[c]{33.25}\\

        \makecell[l]{SAFDNet\dag \cite{zhang2024safdnet}}
        &\makecell[c]{10000}
        &\makecell[c]{15.7M}
        &\makecell[c]{71.0}
        &\makecell[c]{66.3}
        &\makecell[c]{0.03s/0.8G}
        &\makecell[c]{34.68}\\

        \makecell[l]{{FSHNet}\dag \cite{liu2025fshnet}}
        &\makecell[c]{10000}
        &\makecell[c]{11.1M}
        &\makecell[c]{71.7}
        &\makecell[c]{68.1}
        &\makecell[c]{0.03s/0.7G}
        &\makecell[c]{34.17}\\
        
        \Xhline{1pt}
	\end{tabular}
 }
	\caption{The TBK results of multiple detectors for T2LDM++ on text-guided nuScenes. \dag means a fully sparse detector. 'IMT' and 'IMM' indicate \textit{Inference Mean Time} and  \textit{Mean Memory} for each point cloud (32768 points). Run on an RTX 4090 GPU with batch size=1.}
	\label{tab655}
    \vspace{-30pt}
\end{table}


\section{Conclusion}\label{sec7}

In this paper, we proposed a Text-to-LiDAR Diffusion Model, T2LDM++, extended from the previous version \cite{qu2026self}. In this extended version, we provided a deeper analysis of the effective mechanism behind SCRG, redesigning the framework to further reduce computational overhead while maintaining effectiveness. Meanwhile, two high-quality Text-LiDAR benchmarks were constructed, demonstrating the generalization of constructing text descriptions from geometric annotations. Furthermore, we also extended T2LDM++ to multiple conditional generation tasks through a non-latent ControlNet framework, including (Semantic, Box, BEV and Camera)-to-LiDAR, Sparse-to-Dense, and Dense-to-Sparse generation. Moreover, benefiting from the strong correspondence between text descriptions and geometric annotations, we achieved Zero-Shot Text-to-LiDAR generation using Box-to-LiDAR and Semantic-to-LiDAR T2LDM++. In addition, the ablation study on TBK verified the applicability of this controllability metric for generation tasks. \textbf{\textit{Overall, compared with the previous version \cite{qu2026self}, we provided the deeper theoretical insights, the more efficient framework design, and the more comprehensive experimental extensions.}} We hope this work encourages future research toward unified 3D scene understanding and generation through geometry-aware diffusion models.

\section{Limitations}
\label{sec8}

Although T2LDM++ achieves promising performance in both unconditional and conditional LiDAR scene generation, several limitations remain.

\textbf{Limited Generalization for Text Annotation.} Although producing LiDAR scene descriptions from geometric priors is effective, most existing LiDAR datasets (e.g., KITTI-360) lack geometric annotations, limiting the generalization of this strategy. Moreover, the text descriptions are automatically generated from geometric annotations rather than collected from natural language, resulting in limited semantic diversity compared with large-scale Text–Image datasets. Future work may explore alternative annotation priors and more natural semantic rules for scene description generation.

\textbf{Lack of Multi-Conditional Control.} T2LDM++ achieves effective controllable generation under a single condition. However, existing LiDAR datasets usually contain multiple annotation priors, such as 3D boxes and semantic maps. Exploring collaborative control under multiple conditions while exploiting the complementary information of different geometric priors remains an important direction for future research.

\textbf{Reduced Controllability under Incomplete Conditions.} For ControlNet, we observe that controllability is positively correlated with the coverage of input conditions over the generation target. When only partial scene information is available (e.g., Camera-to-LiDAR generation), controllability may degrade. We believe future work can explore combining geometric prior learning with scene completion to recover missing geometric constraints and improve robustness under incomplete conditions.


\section*{Statements and Declarations}

\subsection*{Funding}
This work was supported in part by the Frontier Technologies R$\&$D Program of Jiangsu under grant BF2024070, in part by the National Natural Science Foundation of China under Grant 62471235 and in part by Inspur Storage Qinglan Foundation and Shandong Information Storage System Technology Innovation Center.

\subsection*{Competing Interests}

The authors have no relevant financial or non-financial interests to disclose.

\subsection*{Code Availability}
The source code is publicly available at \href{https://github.com/QWTforGithub/T2LDM_v2}{https://github.com/QWTforGithub/T2LDM$\_$v2}.

\subsection*{Data Availability Statement}
The KITTI-360 dataset \cite{liao2022kitti}  is available at \href{https://www.cvlibs.net/datasets/kitti-360/download.php}{https://www.cvlibs.net/datasets/kitti-360}. The SemanticKITTI dataset \cite{behley2019semantickitti} is available at \href{https://semantic-kitti.org/}{https://semantic-kitti.org/}. The nuScenes dataset \cite{caesar2020nuscenes} is available at \href{https://www.nuscenes.org/nuscenes}{https://www.nuscenes.org/nuscenes}. The T2nuScenes++ (Sec.~\ref{sec51}) and T2SemanticKITTI (Sec.~\ref{sec52}) datasets are available at \href{https://github.com/QWTforGithub/T2LDM_v2}{https://github.com/QWTforGithub/T2LDM$\_$v2}.

\begin{appendices}

\section{}
\label{app_sec1}

\subsection{Texts of T2nuScenes++}
\label{app_sec11}

T2nuScenes++ contains 150,883 Text-to-LiDAR pairs with 65 different text description types.

\noindent  \textbf{All of Classes.}

\noindent barrier : 10709

\noindent bicycle : 7017

\noindent bus : 10533

\noindent car : 131883

\noindent vehicle : 7861

\noindent motorcycle : 7266

\noindent pedestrian : 26397

\noindent traffic cone : 13294

\noindent trailer : 8380

\noindent truck : 23563

\noindent  \textbf{All of Texts.}

\noindent Two cars. : 2099 

\noindent Cars and pedestrians. : 4000

\noindent Cars and traffic cones. : 2000

\noindent Cars and trucks. : 4000

\noindent Cars and vehicles. : 1500

\noindent One car. : 1873

\noindent No cars. : 857

\noindent There are no cars. : 857

\noindent A scene with no cars. : 857

\noindent More than five cars. : 3458

\noindent Cars and barriers. : 1500

\noindent Cars and motorcycles. : 1500

\noindent Five cars. : 1653

\noindent Cars and bicycles. : 1500

\noindent Four cars. : 1842

\noindent Three cars. : 2368

\noindent Cars and buses. : 1500

\noindent Cars and trailers. : 1500

\noindent The scene contains a car and a pedestrian. : 4000

\noindent The scene contains a car and a barrier. : 1500

\noindent The scene contains a car and a traffic cone. : 2000

\noindent The scene contains a car and a vehicle. : 1500

\noindent The scene contains a car and a truck. : 4000

\noindent The scene contains a car and a bus. : 1500

\noindent The scene contains a car and a bicycle. : 1500

\noindent One car is facing backward. : 2487

\noindent One car is facing left. : 5542

\noindent One car is facing right. : 3873
\noindent The scene contains a car and a motorcycle. : 1500

\noindent One car is facing forward. : 2427

\noindent The scene contains a car and a trailer. : 955

\noindent One car is the left of one barrier. : 3072

\noindent One car is the left of one traffic cone. : 3709

\noindent One car is the right of one traffic cone. : 2653

\noindent One car is the right of one barrier. : 2128

\noindent One car is the right of one pedestrian. : 6114

\noindent One car is the left of one pedestrian. : 7721

\noindent One car is the left of one bus. : 3072

\noindent One car is the right of one bus. : 2741

\noindent One car is the right of one truck. : 4651

\noindent One car is the right of one bicycle. : 1783

\noindent One car is the left of one truck. : 5432

\noindent One car is the left of one bicycle. : 2234

\noindent Rainy. One car is facing backward. : 1050

\noindent Rainy. One car is the right of one pedestrian. : 1843

\noindent Rainy. One car is the right of one traffic cone. : 1409

\noindent Rainy. One car is the left of one truck. : 2797

\noindent Rainy. One car is the left of one traffic cone. : 1523

\noindent Rainy. One car is facing left. : 2096

\noindent Rainy. One car is the right of one truck. : 2683

\noindent Rainy. One car is facing forward. : 826

\noindent Rainy. One car is facing right. : 1698

\noindent Rainy. The scene contains a car and a trailer. : 545

\noindent Rainy. One car is the right of one barrier. : 1310

\noindent Rainy. One car is the left of one pedestrian. : 2719

\noindent Rainy. One car is the left of one bus. : 1034

\noindent Rainy. One car is the right of one bus. : 686

\noindent Rainy. One car is the left of one barrier. : 1199

\noindent One car is the right of one vehicle. : 2372

\noindent One car is the left of one vehicle. : 2489

\noindent One car is the left of one trailer. : 1895

\noindent One car is the right of one trailer. : 1543

\noindent One car is the left of one motorcycle. : 2418

\noindent Rainy. One car is the right of one trailer. : 769

\noindent Rainy. One car is the left of one trailer. : 1173

\noindent One car is the right of one motorcycle. : 1848

\subsection{Texts of T2SemanticKITTI}
\label{app_sec12}

T2SemanticKITTI contains 127140 Text-to-LiDAR pairs with 50 different text description types.

\noindent \textbf{All of Classes.}

\noindent motorcyclist : 2507

\noindent bicyclist : 3315

\noindent motorcycle : 3748

\noindent other vehicle : 8271

\noindent bicycle : 7965

\noindent person : 8235

\noindent pole : 11936

\noindent traffic sign : 11966

\noindent trunk : 11219

\noindent car : 29767

\noindent \textbf{All of Texts.}

\noindent The scene contains motorcyclists and cars.: 635

\noindent Motorcyclists and trunks.: 620

\noindent The scene contains motorcyclists and trafficsigns.: 434

\noindent Motorcyclists and poles.: 538

\noindent A road scene with motorcyclists.: 719

\noindent Motorcyclists in a scene with vegetation.: 719

\noindent There are motorcyclists in the scene.: 719

\noindent Motorcyclists.: 719

\noindent A road scene with poles.: 5740

\noindent A road scene with trunks.: 5301

\noindent There are cars in the scene.: 5542

\noindent Poles.: 5642

\noindent There are trunks in the scene.: 5298

\noindent Cars.: 5347

\noindent Traffic signs.: 4032

\noindent Trunks.: 5255

\noindent Cars in a scene with vegetation.: 5495

\noindent There are poles in the scene.: 5658

\noindent Traffic signs in a scene with vegetation.: 2500

\noindent Trunks in a scene with vegetation.: 5208

\noindent There are traffic signs in the scene.: 2500

\noindent A road scene with cars.: 5471

\noindent A road scene with traffic signs.: 2500

\noindent There are persons in the scene.: 1715

\noindent The scene contains other vehicles and cars.: 2000

\noindent Other vehicles in a scene with vegetation.: 1500

\noindent Other vehicles.: 1771

\noindent The scene contains persons and cars.: 4812

\noindent Persons.: 1684

\noindent A road scene with persons.: 1708

\noindent Persons in a scene with vegetation.: 1744

\noindent There are other vehicles in the scene.: 1500

\noindent A road scene with other vehicles.: 1500

\noindent Poles in a scene with vegetation.: 5648

\noindent The scene contains bicyclists and cars.: 1099

\noindent Bicyclists in a scene with vegetation.: 1105

\noindent A road scene with bicyclists.: 1095

\noindent Bicyclists and cars.: 1130

\noindent There are bicyclists in the scene.: 1121

\noindent Bicyclists.: 1139

\noindent The scene contains bicycles and cars.: 5301

\noindent Bicycles in a scene with vegetation.: 1327

\noindent There are bicycles in the scene.: 1416

\noindent A road scene with bicycles.: 1248
Bicycles.: 1388

\noindent The scene contains motorcycles and cars.: 1890

\noindent A road scene with motorcycles.: 990

\noindent Motorcycles and cars.: 1887

\noindent Motorcycles.: 966

\noindent Motorcycles in a scene with vegetation.: 996

\noindent There are motorcycles in the scene.: 868

\end{appendices}

\bibliography{sn-bibliography}

\end{document}